# Learning Resilient Elections with Adversarial GNNs

**Hao Xiang Li**[1*] **Yash Shah**[1*] **Lorenzo Giusti**[2]

[1]Department of Computer Science, University of Cambridge, Cambridge, United Kingdom
[2]CERN, Geneva, Switzerland

hxl23@cst.cam.ac.uk ys562@cam.ac.uk lorenzo.giusti@cern.ch

## Abstract

In the face of adverse motives, it is indispensable to achieve a consensus. Elections have been the canonical way by which modern democracy has operated since the 17th century. Nowadays, they regulate markets, provide an engine for modern recommender systems or peer-to-peer networks, and remain the main approach to represent democracy. However, a desirable universal voting rule that satisfies all hypothetical scenarios is still a challenging topic, and the design of these systems is at the forefront of mechanism design research. Automated mechanism design is a promising approach, and recent works have demonstrated that set-invariant architectures are uniquely suited to modelling electoral systems. However, various concerns prevent the direct application to real-world settings, such as robustness to strategic voting. In this paper, **we generalise the expressive capability of learned voting rules, and combine improvements in neural network architecture with adversarial training to improve the resilience of voting rules while maximizing social welfare**. We evaluate the effectiveness of our methods on both synthetic and real-world datasets. Our method resolves critical limitations of prior work regarding learning voting rules by representing elections using bipartite graphs, and learning such voting rules using graph neural networks. We believe this opens new frontiers for applying machine learning to real-world elections.

## 1 Introduction

Although elections are most commonly used to resolve political disagreements, they can be abstracted as a structured tool for resolving disputes and making collective decisions in everyday contexts [1]. They allow groups to **aggregate preferences** so that every stakeholder has a voice and to **ensure fairness** such that each participant gets an equal say. Elections also help to **avoid conflict** by providing a peaceful resolution. Finally, they **legitimize outcomes** because participants are more willing to accept results when everyone had the chance to participate [2].

When a group of voters with individual preferences face the problem of choosing a single candidate among a set of possible outcomes, elections serve as a mechanism to aggregate those preferences and reach a collective decision that reflects the will of the group with applications beyond political systems, such as in multi-agent robotics, decentralized autonomous agents, and recommender systems [3]. In elections, a voting mechanism refers to an algorithmic process to elicit individual preferences, and choose an outcome with certain criteria [4]. A mechanism ideally possesses socially desirable characteristics: the chosen candidate should be preferred over others in a head-to-head comparison [5]; rational voters express their vote truthfully, or at least, honest opinions should not actively harm the chances of their preferred candidate [6]; the process must be fair, and each voter and candidate should have equal representation.

*Equal contribution.

Unfortunately, selecting a deterministic rule that always selects the socially optimal outcome under a set of seemingly reasonable criteria is often an open problem—or worse, it is sometimes provably impossible for such a rule to exist [7], [8]. Significant research has been directed towards finding improved voting mechanisms, leading to an entire field of study known as voting theory, with contributions exploring fairness criteria [8], strategic manipulation [9], and computational complexity [7].

In this work, we study methods to learn undiscovered voting rules. Our main contribution is a welfare-maximising learning methodology which satisfies the voter anonymity, candidate neutrality, and optionally other miscellaneous criteria, generalises to an arbitrary number of candidates, and is robust to strategic voting. To achieve this, we propose: (a) a combination of permutation-equivariant neural networks composed by: a **graph voting network** (GEVN) and a **graph strategy network** (GESN), (b) an algorithmic design of social welfare and monotonicity losses, and (c) adversarial assessment to expect strategic voting.

## 2 Background and Related Works

**Graph Neural Networks:** Let $G = (\mathcal{V}, \mathcal{E})$ be a graph, with $\mathcal{N}_v = \{u \in \mathcal{V} : (u, v) \in \mathcal{E}\}$ being the one-hop neighborhood of node $v$, having neighborhood features $\boldsymbol{X}_{\mathcal{N}_v} = \{\{\boldsymbol{x}_u : u \in \mathcal{N}_v\}\}$, where $\{\{\cdot\}\}$ denotes a multi-set. We define the message passing function, a permutation-invariant function over the $\boldsymbol{X}_{\mathcal{N}_v}$, as:

$$f\left(\boldsymbol{x}_v, \boldsymbol{X}_{\mathcal{N}_v}\right) = \phi\left(\boldsymbol{x}_v, \bigoplus_{u \in \mathcal{N}_v} \psi(\boldsymbol{x}_v, \boldsymbol{x}_u)\right) \tag{1}$$

where $\psi$ and $\phi$ are learnable message, and update functions, respectively, while $\oplus$ is a permutation-invariant aggregation function (e.g., sum, mean, max). A message passing neural network (MPNN) consists of sequentially applied message passing layers of the form specified in Eq. (1).

**Voting Theory:** Consider a set of $n$ voters $V = \{v_1, v_2, ..., v_n\}$ and $m$ candidates $C = \{c_1, c_2, ..., c_m\}$. Let the $n \times m$ utility profile matrix $\boldsymbol{U} \in \mathbb{R}^{n \times m}$ specify ground truth voter preferences, where utility $u_{ij}$ represents how much a voter $v_i$ prefers candidate $c_j$. The social welfare function $\mathrm{sw}_{\boldsymbol{U}}(c \in C) : C \to \mathbb{R}$ denotes the desirability of candidate $c$ under utility profile $\boldsymbol{U}$. For example, utilitarian welfare [10] corresponds to the sum: $\mathrm{sw}_{\boldsymbol{U}}^{\mathrm{util}}(c_j) = \sum_{i \le n} \boldsymbol{U}_{ij}$. Nash welfare [11] is the product: $\mathrm{sw}_{\boldsymbol{U}}^{\mathrm{nash}}(c_j) = \prod_{i \le n} \boldsymbol{U}_{ij}$, and Rawlsian welfare [12] the minimum: $\mathrm{sw}_{\boldsymbol{U}}^{\mathrm{rawl}}(c_j) = \min_{i \le n} \boldsymbol{U}_{ij}$.

**Mechanism Design:** A voting mechanism is a function from a set of elicited preferences from all voters to a single candidate. A probabilistic social choice function (PSCF) extends the voting mechanism definition to output a probability distribution over candidates instead of a single candidate [13]. The social welfare function represents the maximization objective, and is selected by the domain expert of the system. In general, voting rules can be categorised based on the type of information collected. On one hand, for rankings $R = \{1, ..., m\}$, ranked voting rules expect a bijection of candidates to rankings from each voter: $\sigma_i^{\mathrm{rank}} : C \to R$ representing a total order of preferences. On the other hand, cardinal voting rules receive a graded score for each candidate from voters: $\sigma_i^{\mathrm{cardinal}} : C \to \mathbb{R}$, thus allowing voters to express the strength of their support [4], [14], [15].

Our work primarily focuses on learning cardinal voting rules, and *all methods described can be adapted to ranked voting* (Section C) algorithms by taking normalised Borda [16] as voter scores. For a candidate ranked at position $r$, let its Borda score be $1 - \frac{r}{m}$. We also focus on PSCF: a natural choice for learning by neural networks as it leads to differentiable losses and simplifies the design by avoiding ties [13]. Furthermore, PSCF can be converted into a deterministic voting mechanism by taking the argmax over candidate probabilities.

**Related Works:** Aside from maximising a chosen social welfare function, voting theory literature provides an intuitive set of criteria for evaluating voting mechanisms [17]. For example, the Condorcet criterion always selects the candidate with the most pairwise victories, and the monotonicity criterion captures the concept that increasing support for a candidate should be beneficial for the candidate; it is well known the STV mechanism does not fulfill this requirement [7].

The anonymity criterion states that the voting mechanism does not discriminate against participants (i.e., anonymous voting mechanisms are set-input functions). Anil and Bao [2] show that permutation-invariant networks (PIN) are universal approximators of set-input functions. Therefore, DeepSets [18] and Set Transformers [19] are suited to learning voting mechanisms with guaranteed anonymity and impressive empirical results. However, PINs do not generalise to an unseen number of candidates. A similar, but distinct, criterion is the neutrality criterion; the voting mechanism should not bias towards a particular candidate. A neutral voting mechanism should be permutation-equivariant with respect to the candidates; PINs do not guarantee this property.

Unfortunately, Gibbard's theorem [8] and Hylland's theorem [20] state that all reasonable voting mechanisms satisfying desirable criteria can be manipulated by a voter. Informally, for any voting mechanism, either: (a) it is a dictatorship rule where a single voter decides the result; (b) there are only two candidates; (c) the mechanism is not strategy-proof. Remarkably, there is no dominant strategy for any voter. Thus, the optimal ballot for a voter is conditional on other voters in a system.

With a focus on the third case, as voting mechanism designers, we need to study strategic voting; we cannot expect voters to directly report their utility profiles, but must instead expect 'manipulated' votes, which lead to personal gain at social cost. For example, in a cardinal system optimized for Rawlsian welfare, a voter may report the maximum utility for their favourite candidate, and 0 for all other candidates, thus ensuring only their favorite candidate is selected. Please note that the voting mechanism's manipulability depends on the amount of information a voter has [21].

Automated mechanism design is a promising research direction to learn existing and new voting mechanisms. Procaccia et al. [22] proved that traditional voting rules are probably approximately correct learnable (PAC) with neural networks, providing a secure foundation for ML voting mechanisms. Firebanks [23] experiment with learning modified voting mechanisms with soft constraints. Matone et al. [13] use permutation-invariant embeddings of preference profiles and an MLP architecture to learn voting rules that satisfy the participation criterion [6]. Holliday et al. [21] evaluated the robustness of traditional voting rules to manipulation. Recently, Anil and Bao [2] evaluate permutation-invariant architectures as an alternative to simpler multilayer perceptron (MLP) [24] models, thus ensuring voter anonymity. Their method effectively generalises to different voter numbers and utility distributions, and can directly learn welfare-maximising voting mechanisms with behavioral cloning. They show that neural networks can effectively manipulate existing voting rules even when assuming limited information. However, in realistic settings, *the number of candidates may be dynamic, and some voters may exhibit adversarial behaviour*. Existing methods cannot adapt to these cases. Our contribution bridges the gap between machine learning and real-world applications by introducing a more efficient architecture and a welfare-maximizing loss function.

# 3 Learning Resilient Voting Rules via Graph Neural Networks

We are interested in learning resilient mechanisms that optimize for social welfare. In particular, the mechanism should (1) expect strategic voting from the electorate, and (2) be constrained by desirable design criteria. We consider anonymity and neutrality (fairness), and suggest how to constrain for the Condorcet and monotonicity criteria. Furthermore, our proposed method is generalisable to an unbounded number of candidates, a significant advancement with respect to prior work. An overview of our system can be found in Figure 1 (Left).

**Election Bipartite Graphs (EBGs)**: Given that $n = |V|$, $m = |C|$, there are $n \times m$ voter-candidate mappings to consider. If we naively pass the flattened vector into an MLP, the network is fixed for the number of voters and candidates, and does not guarantee anonymity or neutrality. To partially address this (anonymity and generalisation with respect to the number of voters), prior work uses voter permutation-invariant embeddings [23]. Anil and Bao can be viewed as learning embeddings with PIN architectures [2]. Matone and Firebanks [13] use manual embeddings derived from social choice theory literature, followed by an MLP. However, these methods are unable to account for neutrality and generalisation with respect to the number of candidates.

A neural network to learn PSCF should map from voter preference profiles to a probability distribution over candidates. Therefore, it is crucial to select a suitable representation of voter preference profiles that may be passed into a neural network.

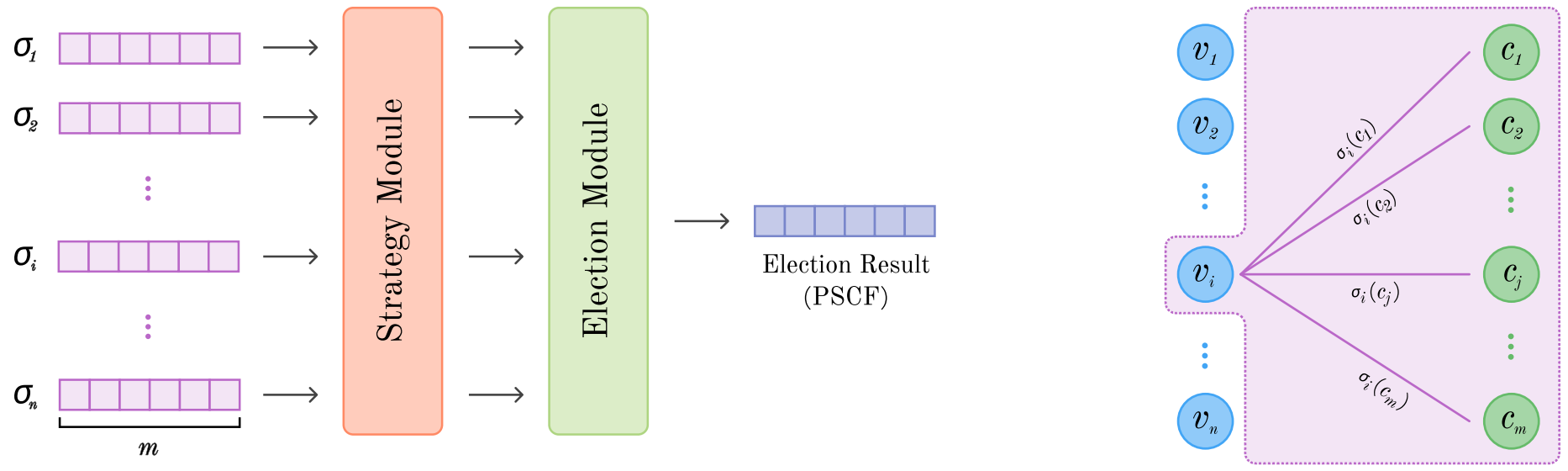

Figure 1: *Left*: An overview of our system. Inputs to the model are the preference profiles of each voter. These are then transformed into strategic preference profiles by the strategy module. Lastly, the election module processes these to produce the result (PSCF). The arrows show data flow. *Right*: An illustration of the $n \times m$ preference profile illustrated as a bipartite graph, with the voters and candidates as nodes, and preference scores as edges. Note that only the edges for voter $i$ have been displayed.

Motivated by Pointer-Networks [25], we represent an election as an attributed undirectional graph, and generalise prior works to an arbitrary number of candidates. Therefore, we propose transforming voters' preferences into an election bipartite graph (EBG), a disjoint set of voter nodes $V$ and candidates $C$. Each node is associated with a one-hot encoding to distinguish between voters and candidates. Every edge bidirectionally connects some voter $v_i$ and some candidate $c_j$, representing the preference $\sigma_i(c_j)$—the one dimensional feature of edge $(v_i, c_j)$ is $x_{v_i c_j} = \sigma_i(c_j)$. Since preference profiles are fully defined across all voter-candidate pairs, the EBG is a complete bipartite graph (illustrated in Figure 1 (Right)).

**Election Module:** To guarantee anonymity and neutrality, it is sufficient to learn permutation-equivariant functions with an EBG as input. Evaluating winner probabilities becomes a node-level classification task across candidate nodes with voter nodes masked.

**Proposition 3.1.** A permutation equivariant function over an EBG transformed into a voting mechanism by masking out voter nodes satisfies anonymity and neutrality.

We provide a proof of the proposition above in Section A.1.

Graph neural networks (GNNs) [26] are a natural choice to learn permutation-equivariant functions on a graph domain. We use a custom graph network (GN) [27], which is an extension of the MPNN to take into account both edge features and node features, and term our model the graph election voting network (GEVN).

Consider a graph $G = (\mathcal{V}, \mathcal{E})$ with nodes $\mathcal{V}$ and edges $\mathcal{E}$. Define features for nodes $x_i$ and edges $x_{ij}$, as well as latents $h_i^l, h_{ij}^l$, where $l$ refers to the layer and initial layer features $h_i^0 = x_i, h_{ij}^0 = x_{ij}$. Each layer of the network updates latent features as follows:

$$\begin{aligned} h_{ij}^{l+1} &= \phi_e\left(h_i^l, h_j^l, h_{ij}^l\right) \\ h_i^{l+1} &= \phi_v\left(h_i^l, \bigoplus_{j \in \mathcal{N}_i} \psi\left(h_i^l, h_j^l, h_{i,j}\right)\right) \end{aligned} \tag{2}$$

$\phi_e, \phi_v, \psi$ are edge operation, node operation, and message passing operations, respectively. We use MLPs as building blocks to define parameterised operations within a GEVN layer. After several such layers, we apply a candidate mask on nodes and use the softmax function to transform scalar values into a probability distribution; thus, the GEVN can be interpreted as a PSCF.

It is desirable for any proposed architecture to be sufficiently powerful to learn arbitrary voting mechanisms. Building upon connections between graph isomorphism testing and universal function approximation by Xu et al. [28] and Chen et al. [29], we provide a proof that the GEVN is sufficiently expressive in Section A.2.

**Theorem 3.2.** A GNN that maps any two graphs that the Weisfeiler-Lehman test decides as non-isomorphic to different embeddings will be a universal approximator when applied to the EBG.

Prior work on training [2], [13], [23] primarily learnt existing voting rules with behavioral cloning; the PSCF is trained to classify the correct candidate by minimizing negative log-likelihood (NLL) loss. Since we are interested in learning efficient voting mechanisms, it is possible to directly optimise for the social welfare function. The expected social welfare of a PSCF $p$ over candidates is:

$$\mathbb{E}[\text{sw}_{\boldsymbol{U}}(p)] = \sum_j p(c_j)\text{sw}_{\boldsymbol{U}}(c_j) \tag{3}$$

We define the welfare loss as

$$L_{\text{sw}_{\boldsymbol{U}}} = -\mathbb{E}[\text{sw}_{\boldsymbol{U}}(p)] \tag{4}$$

Maximising the expected social welfare corresponds to minimising the welfare loss.

**Satisfying Voting Criteria:** The GEVN module can be constrained to fulfill various voting criteria as desired. For example, the Condorcet criterion is guaranteed by truncating the support of the GEVN output to the Smith set [30]. Furthermore, additional criteria may be captured practically using training losses. For instance, notice the monotonicity criterion can be expressed as a first-order constraint $\forall v_i, c_j.\ \frac{\partial p(c_j)}{\partial(\sigma_i(c_j))} \geq 0$ stating a candidate's assignment probability is non-decreasing with respect to the preference expressed for said candidate. Let the monotonicity loss be:

$$L_{\text{mono}} = -\sum_{i,j} \mathbb{1}_{\{\partial p(c_j)/\partial(\sigma_i(c_j))<0\}} \partial p(c_j)/\partial(\sigma_i(c_j)) \tag{5}$$

This is tractable by backpropagating twice through the computation graph.

**Strategy Module:** We define strategic voting as a scenario in which the voter's preference profile is not a true reflection of their underlying utilities. Such behaviour is often observed in the real-world when voters choose to vote tactically by choosing a candidate that they do not have the highest preference for (the candidate that maximises that voter's utility), but has a reasonable probability of winning, given the utility trade-off [31]. A desirable voting mechanism should choose a socially desirable candidate even in the presence of tactical voting.

To learn robust voting mechanisms, we propose to model strategic voter behaviour with a graph election strategy network (GESN) and jointly train the GEVN with GESN in an adversarial manner (See Figure 2).

The GESN maps the utility profile $\boldsymbol{U}$ to a set of preference profiles for each voter. The choice of network architecture can be used to enforce constraints on information available to the strategic voter on a spectrum—from global knowledge of all true utilities, to only personal utilities. Core to our method, the GESN is trained by minimising a rational loss for each voter. A rational voter $v_i$ aims to optimise for its own personal utility by controlling its submitted preference profile as an action:

$$\max_{\sigma_i} \sum_{j=1\ldots m} p(j|\sigma_i)\boldsymbol{U}_{ij} \tag{6}$$

$$L^i_{\text{rational}} = -\sum_{j=1\ldots m} p(j|\sigma_i)\boldsymbol{U}_{ij} \tag{7}$$

To apply necessary restrictions on the action space, when applying gradient descent for the GESN, the weights of the GEVN are frozen. Furthermore, during backpropagation of $L^i_{\text{rational}}$, we cut off gradients at unrelated actions to $v_i$. This rational loss allows the GESN to learn to modify the input preference profile for each voter in a manner that maximises their individual utility. The GESN output is normalised to restrict the preference profile space to a compact space, and captures fairness in the sense of one-person-one-vote. For example, where $a, b \in \mathbb{R}$, we constrain votes to sum to $a$ (budget of votes), or constrain votes to a range of $[a, b]$.

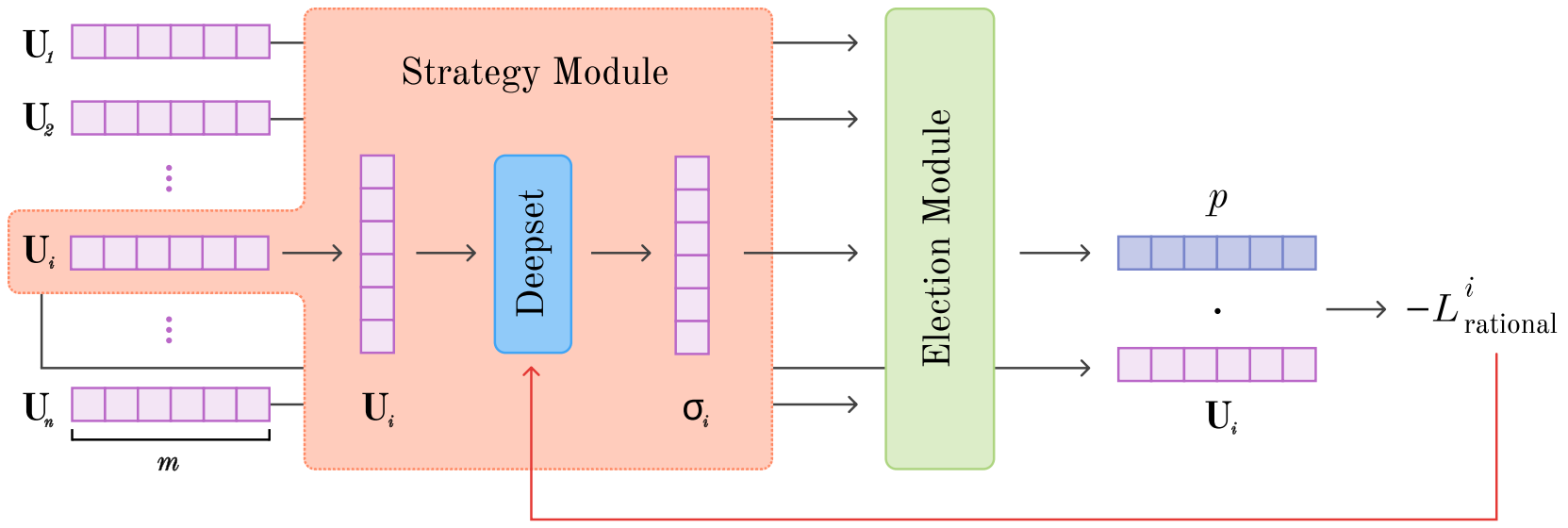

Figure 2: This figure illustrates the method used to train the strategy module, using a rational loss. In this example, voter $v_i$ has been selected as the target voter to optimise the GESN for, and we block the propagation of gradients to the other voters. The arrows show the data flow. This specific architecture assumes private utility information and no communication links between voters, but our general framework is not limited to such assumptions.

The GEVN is trained to optimise the $L_{\text{sw}_U}$ and optionally $L_{\text{mono}}$ while the strategy module optimises for $\sum_i L^i_{\text{rational}}$. Since the GEVN is trained in an end-to-end manner simultaneously with the GESN, it learns to adapt to the voting manipulations made.

# 4 Experiments

In this section, we empirically verify the potential benefits of our model. Precisely, the following questions are of interest. (1) Is the GEVN sufficiently powerful to learn voting mechanisms, and is it able to generalise over a variable number of voters/candidates? (2) Can the GEVN successfully learn welfare-maximising voting mechanisms? (3) Is the voting mechanism learnt during our adversarial training process resilient against strategic voters?

**Datasets:** We sample impartial synthetic profiles [32] using the *Dirichlet* distribution with $\alpha = 1$ and the *spatial* model where voters and candidates are uniformly distributed in a unit cube with utilities $\boldsymbol{U}_{ij} = 1 - \|v_i - c_j\|_2$. We also examine our method using real-world datasets: the *MovieLens* dataset [33] and an experiment during the 2017 French elections, *Grenoble* city [34]. We uniformly subsample elections with the desired number of candidates and voters as needed.

**Training Setup:** We use the Adam optimiser [35] and Cosine annealing learning rate scheduling with warm restarts [36] to train the modules with gradient descent. A complete set of architecture and experimental details is deferred to Section E.

**Learning Classical Voting Rules:** To validate the expressiveness of our election module, we manually calculate prevailing voting rules (Plurality, Borda, Copeland, Maximin, and Single Transferable Vote (STV). See Section B for details.), and then train our model to mimic these classical methods using NLL loss. We generate training data with the number of voters selected from a range of 3 to 50, and the number of candidates from 2 to 10. Furthermore, we use a validation dataset of 75 voters and 15 candidates to select the best model in a training run, to then evaluate on the test set of 100 voters and 20 candidates.

As a comparison, we also implement a DeepSets election model with either Borda scores or one-hot encoding as described in [2] to reflect candidate rankings. In prior work [2], DeepSets were able to achieve state-of-the-art (SOTA) performance. Because DeepSets cannot extrapolate to a larger number of candidates, we revise the validation set to 10 candidates and omit the test set.

As shown in Table 1, our proposed GEVN achieves SOTA performance when tasked with mimicking classical voting rules. In particular, score based classical rules are exceptionally well modelled by GEVN, achieving a top accuracy of 0.92 on plurality and perfect accuracy on Borda in the test set. The generalisation accuracy of GEVN on the validation set greatly exceeds that of the DeepSet, despite being a *strictly harder* task with OOD candidate numbers. This generalisation accuracy carries over to the much harder test set, as well as the real-world datasets (deferred to Section D.1). We conjecture the performance improvement and generalisation capabilities are due to neutrality enforced by permutation equivariant GNNs: all the considered classical rules are neutral, and no model capacity is spent approximating neutrality.

Table 1: Learning classical voting rules with machine learning. The expected accuracy, with its 95% confidence interval across 10 repeats, is displayed. Small models are defined with a parameter budget of $100,000$, otherwise the parameter budget is $1,000,000$. The GEVN substantially outperforms the DeepSet model, achieving SOTA performance. *Omitted confidence intervals are those where the* 95% *interval is below* 0.01.

| | | **GEVN** | **GEVN (Small)** | **DeepS.** | **DeepS. (Small)** | **DeepS. OneHot** | **DeepS. OneHot (Small)** |
|---|---|---|---|---|---|---|---|
| **Plurality** | Validation | $0.99 \pm 0.01$ | $\mathbf{1.00}$ | $0.48 \pm 0.05$ | $0.25 \pm 0.04$ | 0.83 | $0.80 \pm 0.01$ |
| | Test | $\mathbf{0.92} \pm 0.04$ | $0.91 \pm 0.03$ | [-] | [-] | [-] | [-] |
| **Borda** | Validation | $\mathbf{1.00}$ | $\mathbf{1.00}$ | $0.41 \pm 0.05$ | $0.38 \pm 0.05$ | $0.46 \pm 0.01$ | $0.44 \pm 0.01$ |
| | Test | $\mathbf{1.00}$ | $\mathbf{1.00}$ | [-] | [-] | [-] | [-] |
| **Copeland** | Validation | $\mathbf{0.76}$ | $\mathbf{0.76}$ | $0.37 \pm 0.04$ | $0.41 \pm 0.05$ | $0.45 \pm 0.01$ | $0.43 \pm 0.02$ |
| | Test | $\mathbf{0.78}$ | $\mathbf{0.78}$ | [-] | [-] | [-] | [-] |
| **Minimax** | Validation | $\mathbf{0.73}$ | $\mathbf{0.73}$ | $0.40 \pm 0.04$ | $0.45 \pm 0.04$ | $0.43 \pm 0.01$ | $0.40 \pm 0.01$ |
| | Test | $\mathbf{0.70}$ | $\mathbf{0.70}$ | [-] | [-] | [-] | [-] |
| **STV** | Validation | $\mathbf{0.53}$ | $\mathbf{0.53}$ | $0.34 \pm 0.04$ | $0.35 \pm 0.04$ | $0.42 \pm 0.01$ | $0.40 \pm 0.01$ |
| | Test | $\mathbf{0.49}$ | $\mathbf{0.49}$ | [-] | [-] | [-] | [-] |

Furthermore, the GEVN architecture scales well in size, using a constant number of parameters, independent of the number of voters or candidates; in contrast, the number of parameters for a DeepSets model will require increasing the number of parameters for the input and output layer when the number of candidates are increased. We note that a small 100k parameter budget GEVN performs comparably to that of a 1 million parameter model, further demonstrating the efficiency of the GEVN architecture.

**Learning to Maximise Welfare:** Previous work [2], [13], [23] used NLL loss which we refer to as the *rule* loss that facilitates learning voting rules through behavioural cloning. We argue that since the overarching aim of a voting system is to maximise some welfare function over a population of voters, our proposed welfare loss $L_{\text{sw}} = -\mathbb{E}[\text{sw}_{\boldsymbol{U}}(p)]$ that directly optimises for welfare, is a better means to achieve this goal.

Our new welfare loss outperforms the rule loss when the aim is to maximise welfare, and we do not have access to the true underlying voter utilities. We see from the results in Figure 3 (Left) (and additional data in Section D.2) that welfare loss consistently achieves higher welfare across the three welfare functions tested (Utilitarian, Nash, and Rawlsian). We attribute this improvement to the soft penalty structure of welfare loss. Unlike rule loss, which applies an equal penalty for any incorrect prediction, welfare loss differentiates between near-optimal and poor choices.

In Section D.3, we see that rule loss often achieves higher accuracy in finding the winner than welfare loss on the validation set, and occasionally on the test set (Figure 3 (Right)). However, this is inconsequential as we are not aiming to optimise for how well the model can identify the winner (or learn the voting rule/welfare function), but are instead trying to maximise social welfare, which the welfare loss does so successfully.

When the true underlying voter utility data is available, we see that our welfare loss performs almost identically to the rule loss. This is expected, as maximizing welfare is trivial when the true cardinal utilities are known. Data supporting this claim can be found in Section D.4.

Our experiments with learning different welfare functions with monotonicity loss show that GEVN can successfully enforce the monotonicity property. A brief analysis can be found in Section D.5.

**Adversarial Training against Strategic Voting:** We additionally examine if our learnt voting mechanisms can adapt to strategic voting behaviour. Strategic voters are modelled under three increasingly informative priors. In the *private* setting, each strategic voter observes their own utility profile. In the *public* setting, strategic voters observe the utility matrix of the entire electorate. In the *results* setting, strategic voters are given access to the election outcome that would arise if all voters reported truthfully. In all cases, voters are sampled uniformly such that an expected 20% of the electorate use strategic votes produced by GESN, and the remainder report true utilities.

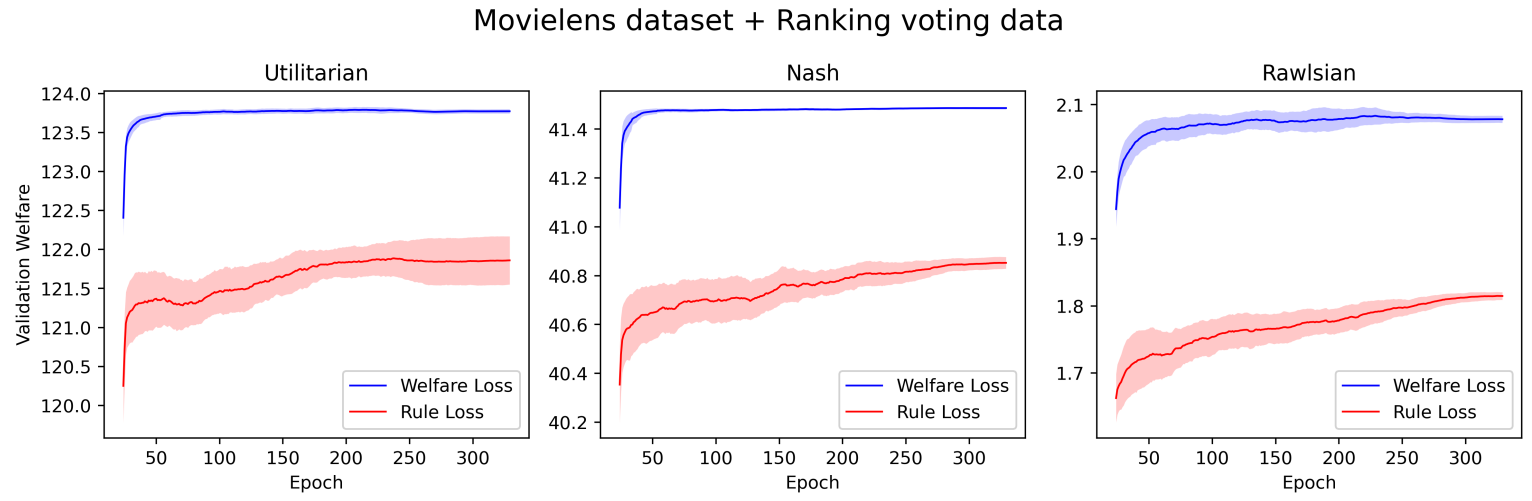

| Acc. | $L_{\text{sw}}$ | $L_{\text{rule}}$ |
|---|---|---|
| Util. | **0.681** ± 0.003 | 0.674 ± 0.009 |
| Nash | 0.649 ± 0.007 | **0.650** ± 0.011 |
| Rawl. | 0.393 ± 0.005 | **0.397** ± 0.010 |

Figure 3: *Left:* The graphs show the validation *welfare* under each welfare function when the GEVN architecture is trained with our proposed welfare loss (blue line) and the rule loss (red line), with voters' candidate rankings as input. The darker line indicates the mean, with the shaded region showing one standard deviation above and below the mean, across five runs. *Right:* The table shows the mean and standard deviation across the same five runs for *accuracy* over the test set, with the best result in **bold**. $L_{\text{sw}}$ refers to (social) welfare loss and $L_{\text{rule}}$ refers to rule loss.

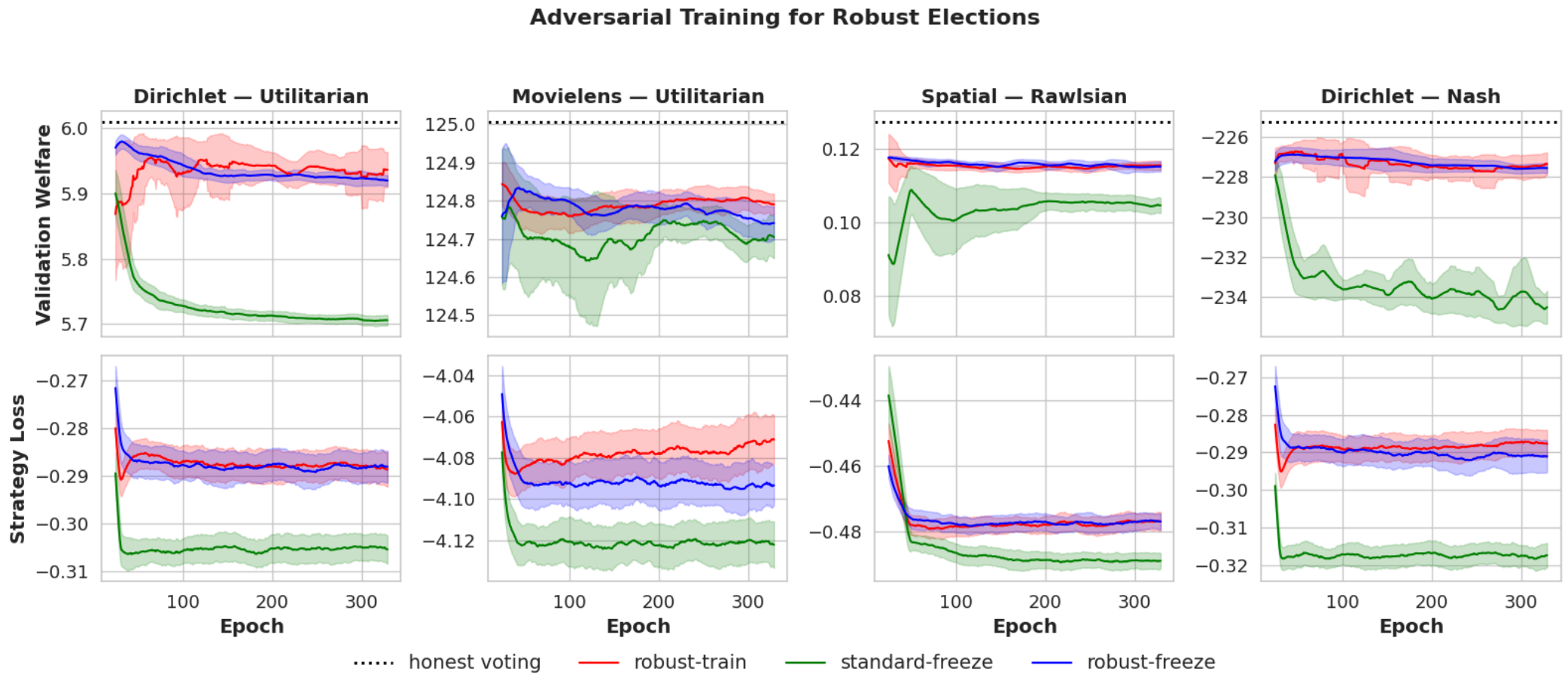

Figure 4: Jointly training the GEVN and GESN modules (*Private* information). *Top*: Social welfare on the validation dataset *Bottom*: Rational loss during training. **standard-freeze**: Freezing the weights of a pretrained GEVN on truthful votes, and training the GESN. **robust-train**: Jointly training the GESN and fine-tuning the pretrained honest GEVN. **robust-freeze**: Taking the output of robust-train as the pretrained model, and training the GESN. The graphs are smoothed using a moving average over 25 epochs, and we plot the 95% confidence interval of the mean.

Our results for the *private* information setting are shown in Figure 4, with the *public* and *result* information setting reported in Section D.6. Across all objectives, strategic voting leads to lower welfare. At the same time, the rational loss decreases, suggesting that strategic voters were able to extract a measurable advantage by not voting honestly even with limited information [21].

In the *private* and *public* information settings, the standard-frozen case has significantly lower social welfare relative to the two robust scenarios. This demonstrates that the proposed adversarial training process improves resilience to a proportion of strategic voting. In particular, we would like to highlight the robust-freeze scenario, which only trains the GESN against a fixed, pre-trained robust GEVN module. In many scenarios, including all *private* information settings, a new set of strategic voters was not able to achieve a better outcome compared to the GESN used in joint training.

In constrast, the robust-freeze scenario does not consistently succeed where voters have access to the truthful election outcome. In this regime, the trained GEVN remains partially exploitable (Figure 8). This reflects the inherent difficulty of learning robust election rules against a population of suffi-ciently informed and motivated actors, and the theoretical constraints implied by Gibbard's theorem. To our knowledge, our work constitutes the first systematic exploration of adversarially trained GNNs in general elections, and the resilient performance in the remaining robust-train settings demonstrate future potential of learnt resilience with real-world assumptions.

## 5 Limitations, Discussions, and Conclusions

In this work, we propose a method to learn robust voting mechanisms with graph neural networks by representing elections using bipartite graphs, ensuring voter anonymity, candidate neutrality, and generalisation to arbitrary numbers of voters and candidates.

Our code is open-source, and available at https://github.com/MarkHaoxiang/geometric-governance.

**Limitations and Future Work:** Our strategy module instantiations considers an incomplete set of information priors which may not match real-world behaviour. Augumenting the framework with other classes of cross-voter signals offers a clear path forward. In addition, because we analyse only a single election, we do not capture the richer dynamics that emerge over sequential elections, where repeated interaction can unlock cooperative equilibria (e.g., the classic prisoner's-dilemma solution) and better mirror real-world behaviour. Computational constraints meant we could not fully explore adversarial training varying proportions of strategic agents—an empirical space we intend to chart once larger budgets become available. Finally, an avenue of future work is exploring additional constraints such as consistency: merging two electorates with the same winner preserves the same winner. Further research into meeting various constraints, particularly via geometric or structural methods, can help establish theoretical worst-case guarantees on the learnt voting rules and enhance user trust.

**Conclusions:** We propose a practical method to learn robust voting mechanisms with graph neural networks. We first represent an election using the EBG and train a voting mechanism with the GEVN, which ensures voter anonymity, candidate neutrality, and generalisation in both voter number and candidate number by design. It is also a universal approximator over the design space of voting mechanisms. Then, we propose the welfare loss to improve unsupervised learning of social welfare maximising mechanisms, and the monotonicity loss to respect the monotonicity constraint. Lastly, we model the voting patterns of strategic voters using the GESN and train adversarially with the GEVN to improve resilience.

**Broader Impact:** Our method achieves empirically impressive performance learning both existing voting rules and new voting rules, while being resilient to rational voting patterns. These contributions provide solutions to key problems impeding prior work from real-world application. We are confident that this method will unlock new directions for applying machine-learning techniques to real-world electoral contexts.

# A Proofs

## A.1 Proof of Anonymity and Neutrality

Let $G = (V \cup C, E)$ be an Election Bipartite Graph (EBG), where $V = (v_1, ..., v_n)$ is the set of voters, and $C = (c_1, ..., c_m)$ is the set of candidates. The topology of $G$ is fully encoded by the adjacency matrix $\boldsymbol{A} \in \{0,1\}^{(n+m)\times(n+m)}$. Also, let $\boldsymbol{X} \in \mathbb{R}^{(n+m)\times d}$ be the node feature matrix which contains one-hot encoded vectors to distinguish between voters and candidates such that the first $n$ rows correspond to voters, while the last $m$ rows correspond to candidates.

A function $f : (\boldsymbol{X}, \boldsymbol{A}) \to \mathbb{R}^{n+m}$ is said to be permutation invariant with respect to a symmetry group $S_{n+m}$ iff $\forall \pi \in S_{n+m}$ it holds:

$$f(\boldsymbol{P}_\pi \boldsymbol{X}, \boldsymbol{P}_\pi \boldsymbol{A} \boldsymbol{P}_\pi^T) = f(\boldsymbol{X}, \boldsymbol{A}), \tag{8}$$

where $\boldsymbol{P}_{\{\pi\}}$ is a permutation matrix that permutes the rows and the columns of both $\boldsymbol{X}$ and $\boldsymbol{A}$ according to $\pi$. Without loss of generality, a function $f : (\boldsymbol{X}, \boldsymbol{A}) \to \mathbb{R}^{n+m}$ is said to be permutation equivariant with respect to $S_{n+m}$ iff $\forall \pi \in S_{n+m}$ it holds:

$$f(\boldsymbol{P}_\pi \boldsymbol{X}, \boldsymbol{P}_\pi \boldsymbol{A} \boldsymbol{P}_\pi^T) = \boldsymbol{P}_\pi f(\boldsymbol{X}, \boldsymbol{A}). \tag{9}$$

A voting mechanism $\nu : (\boldsymbol{X}, \boldsymbol{A}) \to \mathbb{R}^m$ isolates the $m$ candidates by masking the result of $f$ with a matrix $\boldsymbol{M} = [\boldsymbol{0}_{m\times n} \mid \boldsymbol{I}_m] \in \{0,1\}^{m\times(n+m)}$. Formally:

$$\nu(\boldsymbol{X}, \boldsymbol{A}) = \boldsymbol{M} f(\boldsymbol{X}, \boldsymbol{A}). \tag{10}$$

Therefore, $\nu$ satisfies **anonymity** iff $\nu$ is *invariant* under any permutation of voters, and **neutrality** iff $\nu$ is *equivariant* under any permutation of candidates [37] .

*Proof:* Let $f$ be a graph neural network (GNN) such that it produces an output $\boldsymbol{z} = \left[\boldsymbol{z}_V^T, \boldsymbol{z}_C^T\right]^T \in \mathbb{R}^{n+m}$, where $\boldsymbol{z}_V \in \mathbb{R}^n$ are the scores for voters $V$ and $\boldsymbol{z}_C \in \mathbb{R}^m$ are the scores for the candidates $C$.

**Proof of Anonymity:**

*Proof.* Let $\sigma \in S_V \cong S_n$ be a permutation acting on the voter nodes $V$, represented as a block-diagonal matrix $\boldsymbol{P}_\sigma \in \{0,1\}^{(n+m)\times(n+m)}$ as follows:

$$\boldsymbol{P}_\sigma = \begin{pmatrix} \boldsymbol{P}_\sigma^{(V)} & \boldsymbol{0}_{n\times m} \\ \boldsymbol{0}_{m\times n} & \boldsymbol{I}_m \end{pmatrix}, \tag{11}$$

where $\boldsymbol{P}_\sigma^{(V)}$ is the $n \times n$ permutation matrix for $\sigma$ acting on voters' nodes. By the permutation equivariance of $f$, it holds:

$$f(\boldsymbol{P}_\sigma \boldsymbol{X}, \boldsymbol{P}_\sigma \boldsymbol{A} \boldsymbol{P}_\sigma^T) = \boldsymbol{P}_\sigma f(\boldsymbol{X}, \boldsymbol{A}). \tag{12}$$

Now, consider a voting mechanism $\nu$, we have:

$$\nu(\boldsymbol{P}_\sigma \boldsymbol{X}, \boldsymbol{P}_\sigma \boldsymbol{A} \boldsymbol{P}_\sigma^T) = \boldsymbol{M} \boldsymbol{P}_\sigma f(\boldsymbol{X}, \boldsymbol{A}). \tag{13}$$

The product $\boldsymbol{M}\boldsymbol{P}_\sigma$ can be decomposed as:

$$\begin{aligned} \boldsymbol{M}\boldsymbol{P}_\sigma &= [\boldsymbol{0}_{m\times n} \mid \boldsymbol{I}_m] \begin{pmatrix} \boldsymbol{P}_\sigma^{(V)} & \boldsymbol{0}_{n\times m} \\ \boldsymbol{0}_{m\times n} & \boldsymbol{I}_m \end{pmatrix} \\ &= \left[\boldsymbol{0}_{m\times n} \cdot \boldsymbol{P}_\sigma^{(V)} + \boldsymbol{I}_m \cdot \boldsymbol{0}_{m\times n} \mid \boldsymbol{0}_{m\times n} \cdot \boldsymbol{0}_{n\times m} + \boldsymbol{I}_m \cdot \boldsymbol{I}_m\right] \\ &= [\boldsymbol{0}_{m\times n} \mid \boldsymbol{I}_m] \\ &= \boldsymbol{M}. \end{aligned} \tag{14}$$

Therefore,

$$\nu(\boldsymbol{P}_\sigma \boldsymbol{X}, \boldsymbol{P}_\sigma \boldsymbol{A} \boldsymbol{P}_\sigma^T) = \boldsymbol{M} f(\boldsymbol{X}, \boldsymbol{A}) = \nu(\boldsymbol{X}, \boldsymbol{A}). \tag{15}$$

Thus, the voting mechanism $\nu$ is invariant to any permutation of the voters, satisfying anonymity.□

**Proof of neutrality:**

*Proof.* Let $\tau \in S_C \cong S_m$ be a permutation acting on the candidate nodes $C$, represented as a block-diagonal matrix $\boldsymbol{P}_\tau \in \{0,1\}^{(n+m)\times(n+m)}$ as follows:

$$\boldsymbol{P}_\tau = \begin{pmatrix} \boldsymbol{I}_n & \boldsymbol{0}_{n\times m} \\ \boldsymbol{0}_{m\times n} & \boldsymbol{P}_\tau^{(C)} \end{pmatrix}, \tag{16}$$

where $\boldsymbol{P}_\tau^{(C)}$ is the $m \times m$ permutation matrix for $\tau$ acting on candidates' nodes.

By the permutation equivariance of $f$, we have:

$$f(\boldsymbol{P}_\tau \boldsymbol{X}, \boldsymbol{P}_\tau \boldsymbol{A} \boldsymbol{P}_\tau^T) = \boldsymbol{P}_\tau f(\boldsymbol{X}, \boldsymbol{A}). \tag{17}$$

Applying the voting mechanism $\nu$:

$$\nu(\boldsymbol{P}_\tau \boldsymbol{X}, \boldsymbol{P}_\tau \boldsymbol{A} \boldsymbol{P}_\tau^T) = \boldsymbol{M}\boldsymbol{P}_\tau f(\boldsymbol{X}, \boldsymbol{A}). \tag{18}$$

The product $\boldsymbol{M}\boldsymbol{P}_\tau$:

$$\begin{aligned} \boldsymbol{M}\boldsymbol{P}_\tau &= [\boldsymbol{0}_{m\times n} \mid \boldsymbol{I}_m] \begin{pmatrix} \boldsymbol{I}_n & \boldsymbol{0}_{n\times m} \\ \boldsymbol{0}_{m\times n} & \boldsymbol{P}_\tau^{(C)} \end{pmatrix}, \\ &= \left[\boldsymbol{0}_{m\times n} \cdot \boldsymbol{I}_n + \boldsymbol{I}_m \cdot \boldsymbol{0}_{m\times n} \mid \boldsymbol{0}_{m\times n} \cdot \boldsymbol{0}_{n\times m} + \boldsymbol{I}_m \cdot \boldsymbol{P}_\tau^{(C)}\right] \\ &= \left[\boldsymbol{0}_{m\times n} \mid \boldsymbol{P}_\tau^{(C)}\right] \\ &= \left[\boldsymbol{P}_\tau^{(C)} \cdot \boldsymbol{0}_{m\times n} \mid \boldsymbol{P}_\tau^{(C)} \cdot \boldsymbol{I}_m\right] \\ &= \boldsymbol{P}_\tau^{(C)} [\boldsymbol{0}_{m\times n} \mid \boldsymbol{I}_m] \\ &= \boldsymbol{P}_\tau^{(C)} \boldsymbol{M}. \end{aligned} \tag{19}$$

Thus, $\boldsymbol{M}\boldsymbol{P}_\tau = \boldsymbol{P}_\tau^{(C)} \boldsymbol{M}$. Substituting back in Equation (18):

$$\begin{aligned} \nu(\boldsymbol{P}_\tau \boldsymbol{X}, \boldsymbol{P}_\tau \boldsymbol{A} \boldsymbol{P}_\tau^T) &= \boldsymbol{P}_\tau^{(C)} \boldsymbol{M} f(\boldsymbol{X}, \boldsymbol{A}) \\ &= \boldsymbol{P}_\tau^{(C)} \nu(\boldsymbol{X}, \boldsymbol{A}). \end{aligned} \tag{20}$$

Therefore, the scores $\nu(\boldsymbol{X}, \boldsymbol{A})$ for candidates $C$ are permutated according to $\boldsymbol{P}_\tau^{(C)}$, that is the permutation applied to the candidates themselves, yielding neutrality. □

Since $\nu$ is invariant under all voters' permutations (anonymity axiom), and equivariant under all candidate permutations (neutrality axiom), the proposition is satisfied. ■

### A.2 Proof of GEVN Universality

We restate the definition of isomorphism for labelled graphs [38]. Let $l, l'$ represent the labelling function over vertices.

**Definition 1.1.** (Isomorphism of Labeled Graphs) $G = (\mathcal{V}, \mathcal{E}, l)$ is isomorphic to $G' = (\mathcal{V}', \mathcal{E}', l')$, iff there exists a bijective function $f : \mathcal{V} \to \mathcal{V}'$ such that

$$\forall u \in \mathcal{V}. l(u) = l'(f(u)), \text{and} \tag{21}$$

$$\forall u, v \in \mathcal{V}. (u, v) \in \mathcal{E} \leftrightarrow (f(u), f(v)) \in \mathcal{E}' \tag{22}$$

We show that the 1-WL [39] test is complete over voting graphs. Since the 1-WL test cannot handle edge features, we encode edge information as labels over additional nodes.

**Definition 1.2.** (Tripartite EBG) The tripartite EBG (T-EBG) $(V, \mathcal{E}, l)$ is a representation of a EBG $(V \cup C, \mathcal{E}', l')$ with $\mathcal{V} = V \cup C \cup U$. $U$ is the elicited set of preference profiles, $V$ is the set of voters, and $C$ is the set of candidates: $\{u_{vc} \mid \forall (v, c) \in \mathcal{E}'\}$. Next, for each $u_{vc}$, construct $(v, u)$ and $(u, c)$ as two edges in $\mathcal{E}$. As in the EBG, each node in the T-EBG is labelled with its purpose:

$$l(u) = \begin{cases} \text{0'} & \text{if } u \in V \\ l'(v, c) & \text{if } u_{vc} \in U \\ \text{1'} & \text{if } u \in C \end{cases} \tag{23}$$

Note that the T-EBG forms a tripartite graph, with all edges between $V, U$ or $U, C$.

Recall that the 1-WL algorithm assigns each vertex an initial colour encoding $l$, then iteratively refines colours with $\text{HASH}(\text{col}(v), \{\text{col}(u) : u \in \mathcal{N}_v\})$. Then, it *fails* to distinguish two graphs $G$ and $G'$ iff, after reaching a stable coloring, the multi-sets of vertex-colours match.

**Lemma 1.3.** Let $G$ and $G'$ be T-EBGs produced from two EBGs as in Definition 1.2. If 1-WL cannot distinguish $G$ and $G'$ then there exists a label preserving isomorphism $f : G \leftrightarrow G'$.

*Proof of Lemma 1.3.* Assume the multisets of stable 1-WL colours coincide, $\{\{\varphi(x) : x \in G\}\} = \{\{\varphi'(x') : x' \in G'\}\}$. We construct a bijection $f : \mathcal{V}_G \leftrightarrow \mathcal{V}_{G'}$ by a **colour-respecting tour** that visits every edge of $G$ exactly once.

1) *Initial match.* Pick an unmatched voter $v \in V_G$ and choose any $v' \in V_{G'}$ with $\varphi(v) = \varphi'(v')$; set $f(v) := v'$. If all vertices are now matched, we are done.

2) *Forward step* $V \leftrightarrow U \leftrightarrow C$. Select an untraversed edge $(v, u) \in \mathcal{E}_G$ with $v$ already matched and $u \in U_G$. Let $(u, c) \in \mathcal{E}_G$ with $c \in C_G$ be its continuation.

   a) If $c$ *already matched*. The path $(f(v) \leftrightarrow u' \leftrightarrow f(c))$ exists in $G'$ because $G'$ is complete bipartite. Match $u$ with this unique $u'$; 1-WL results in $\varphi(u) = \varphi'(u')$.

   b) If $c$ *unmatched*. Take any $u' \in G'.U$ adjacent to $f(v)$ with $\varphi(u) = \varphi'(u')$ and let $(u', c') \in E_{G'}$. Set $f(u) := u'$ and $f(c) := c'$.

   Continue the tour from $c$; if no outgoing untraversed edge remains, jump to any matched vertex that still has one. Such a vertex exists because the T-EBG is Eulerian.

3) *Backward step* $C \leftrightarrow U \leftrightarrow V$. Symmetric to the forward step, starting from a matched candidate and traversing $(c, u)$ then $(u, v)$.

4) *Restart.* If unmatched vertices remain, return to the forward step with an unmatched voter. Finiteness of $G$ guarantees termination after every vertex has been matched and every edge traversed.

The resulting map $f$ is bijective by construction, satisfies $\varphi(x) = \varphi'(f(x))$ (colour equality implies label equality, proving (21)), and preserves adjacency because each image edge is chosen inside $G'$ (proving (22)). Hence $f$ is the desired label-preserving isomorphism. □

## B Additional Details on Classical Voting Rules

### Plurality Voting

In this voting scheme, each voter votes for their favorite candidate, and the winner of the election is the candidate with the most votes.

If the election was represented by a vote matrix $\boldsymbol{V}$ of size $n \times m$ (number of voters by number of candidates), where $\boldsymbol{V}_{i,j}$ represents the ranked position of candidate $j$ in the preference profile of voter $i$, then each candidate is scored by

$$\text{plurality}(\boldsymbol{V}, j) = \sum_{i=1}^{n} \mathbf{1}_{\boldsymbol{V}_{i,j}=1} \tag{24}$$

### Borda voting

Each candidate is given points corresponding to the Borda count

$$\text{borda}(\boldsymbol{V}, j) = \sum_{i=1}^{n} \left(1 - \frac{\boldsymbol{V}_{i,j}}{m}\right) \tag{25}$$

### Copeland

Copeland voting models the $\frac{m(m-1)}{2}$ pairwise sub-elections in any single election. A candidate scores 1 point for winning a pairwise election, $\frac{1}{2}$ for a draw, and 0 otherwise. The overall winner is the candidate with the highest score.

### Maximin

Similar to the Copeland method, Maximin also models the pairwise elections. A candidate is scored by the number of votes it achieves against its worst matchup.

### Single Transferrable Vote (STV)

In the single transferrable vote election, each voter initially casts votes as plurality voting. After votes are counted, while no candidates achieve an absolute majority of votes, the worst performing candidate is eliminated. Voters who previously cast their votes for that candidate, are allowed to recast votes for the remaining candidates. This elimination process repeats until the winner remains.

# C Application to Ordinal Voting Systems

While our primary focus is on cardinal voting rules, or framework readily extends to ordinal (ranked) voting rules. As noted in Section 2, the signal of a known ranking may be converted to normalised Borda scores (Section B) to serve as edge features in the EBG while preserving anonymity and neutrality guarantees. This transformation applied in our experiments learning classicial voting rules.

Extending the strategy module to ordinal systems requires end-to-end differentiability. Work on differentiable ranking operators [40] allow cardinal scores to be transformed into ordinal rankings smoothly, preserving backpropagation. We focus on cardinal systems in the main text because they naturally suit applications with autonomous agents, and to avoid the computational overheads of differentiable ranking (though theoretically efficient, readily-available implementations at the time of writing did not support GPU training).

# D Additional Experimental Results

## D.1 Learning voting rules: generalisation to real-world datasets

We test the trained GEVN models to real-world datasets. Results are show in Table 2. We verify that GEVN is able to effectively generate to real-world datasets when learning existing voting mechanisms. Similarly to [2], we find a higher accuracy when evaluating on real-world data compared to synthetic data.

## D.2 Welfare loss outperforms rule loss with ranking data

The graphs below show additional data concerning welfare maximisation with ranking data. We see that on both remaining datasets, welfare loss outperforms rule loss. The darker line indicates the mean, with the shaded region showing one standard deviation above and below the mean, across five runs, for each plot.

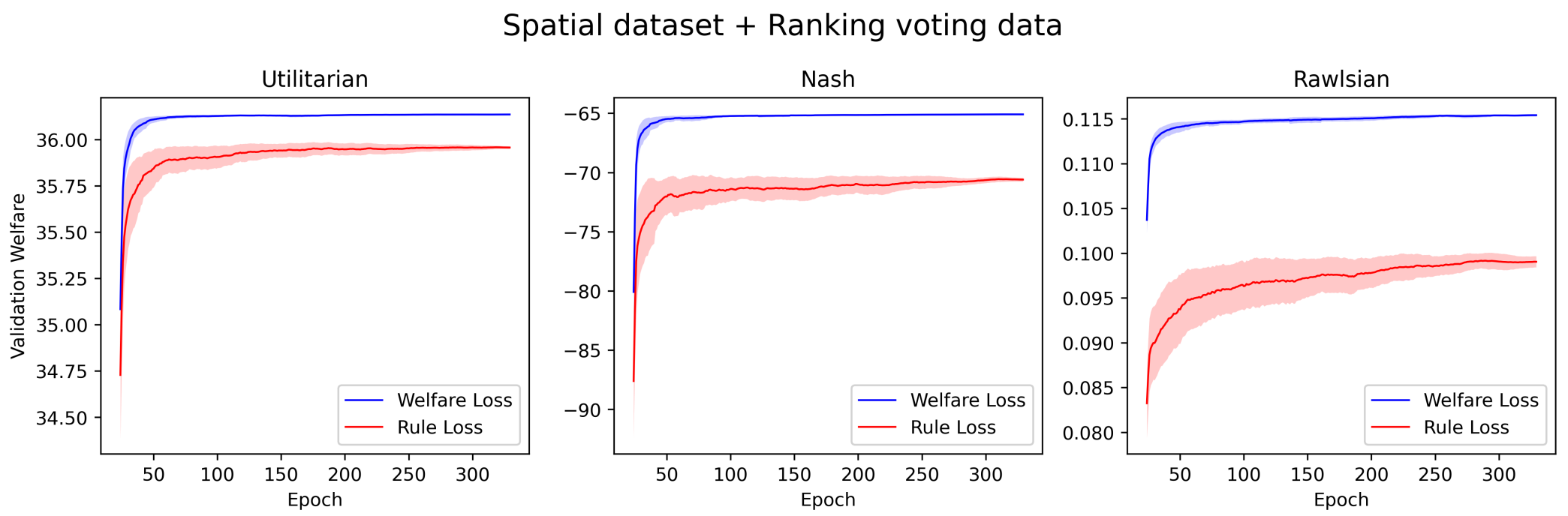

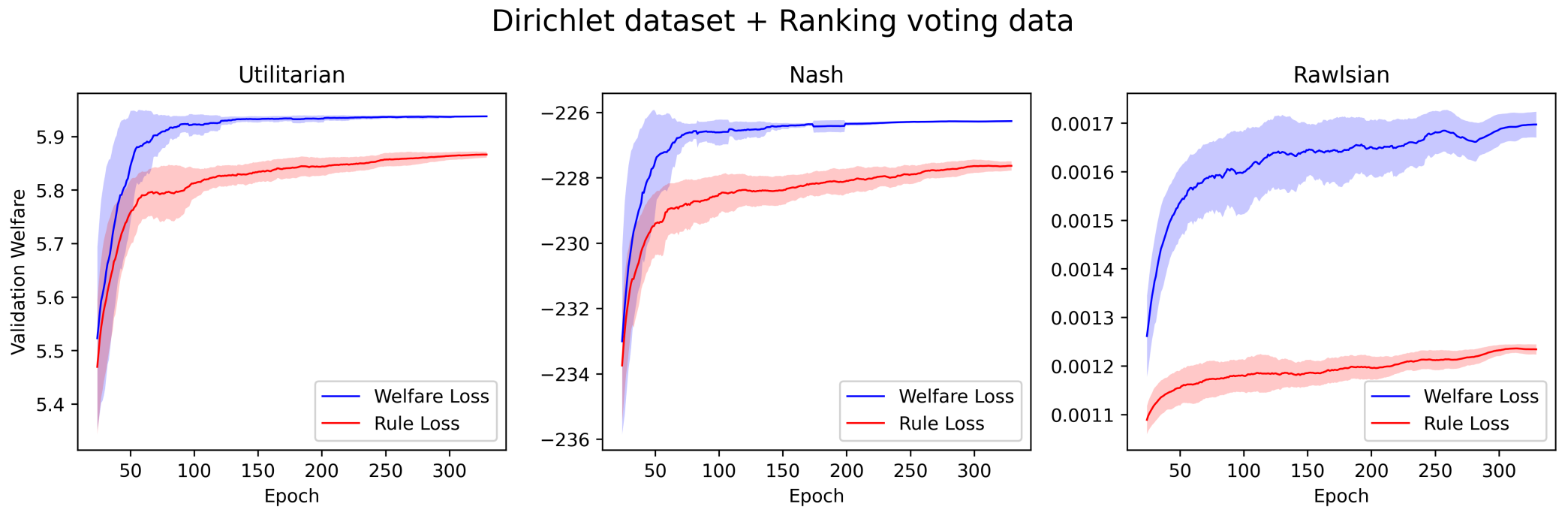

## D.3 Rule loss outperforms welfare loss on accuracy

Table 2: Mimicking classical voting rules tested on real world datasets. For the *Grenoble* dataset, 50 voters and 5 candidates are sampled. For the *Movielens* dataset, 100 voters and 10 candidates are sampled. *Omitted confidence intervals are those where the* 95% *interval is below* 0.01.

| | | **GEVN Small** | **GEVN Medium** |
|---|---|---|---|
| **Plurality** | Grenoble | 1.00 | 1.00 |
| | Movielens | 1.00 | 1.00 |
| **Borda** | Grenoble | 1.00 | 1.00 |
| | Movielens | 1.00 | 1.00 |
| **Copeland** | Grenoble | $0.94 \pm 0.01$ | $0.92 \pm 0.01$ |
| | Movielens | $0.89 \pm 0.01$ | 0.89 |
| **Maximin** | Grenoble | $0.92 \pm 0.01$ | $0.92 \pm 0.01$ |
| | Movielens | $0.90 \pm 0.01$ | $0.90 \pm 0.01$ |
| **STV** | Grenoble | $0.94 \pm 0.01$ | $0.95 \pm 0.01$ |
| | Movielens | 0.87 | 0.87 |

The graphs below show the validation *accuracy* of welfare and rule loss across all three welfare functions (Utilitarian, Nash, and Rawlsian), with both voter rankings and utilities as input. The darker line indicates the mean, with the shaded region showing one standard deviation above and below the mean, across five runs, for each plot.

We often observe that rule loss has better accuracy than welfare loss on ranking data, however this is inconsequential since the primary aim is to maximise wefare. Both loss functions perform similarly when true voter utility data is available. Note that on the validation set for the Dirichlet dataset, convergence was not reached, which we believe could be mitigated with a sufficiently large training set.

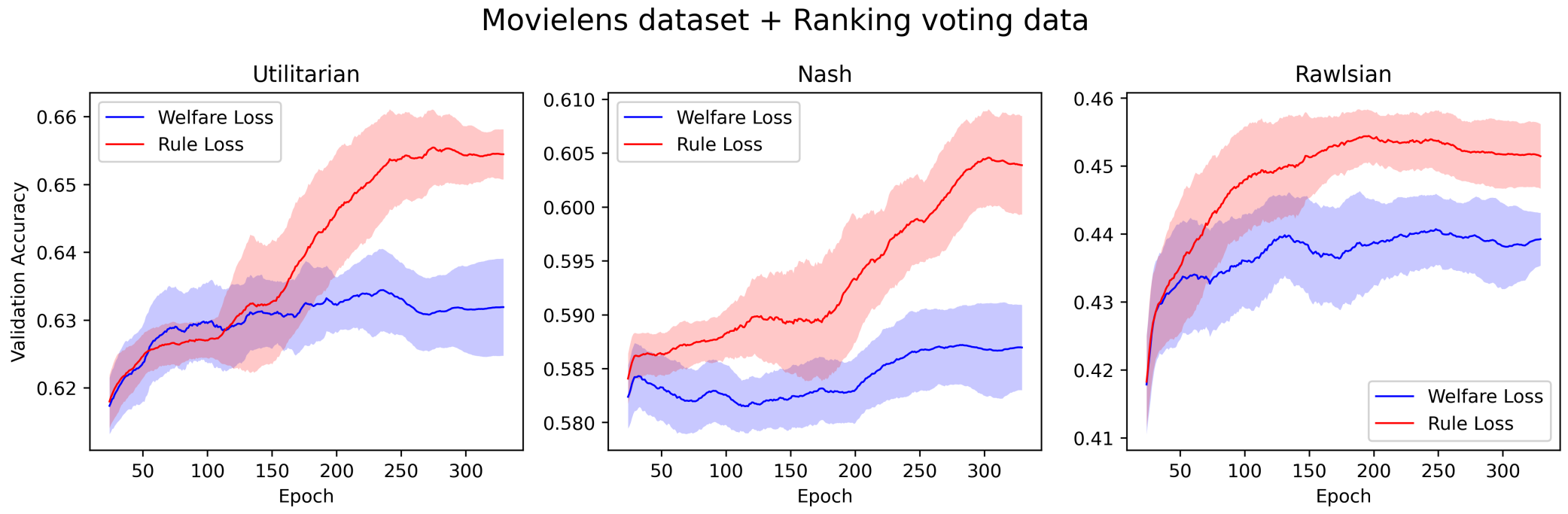

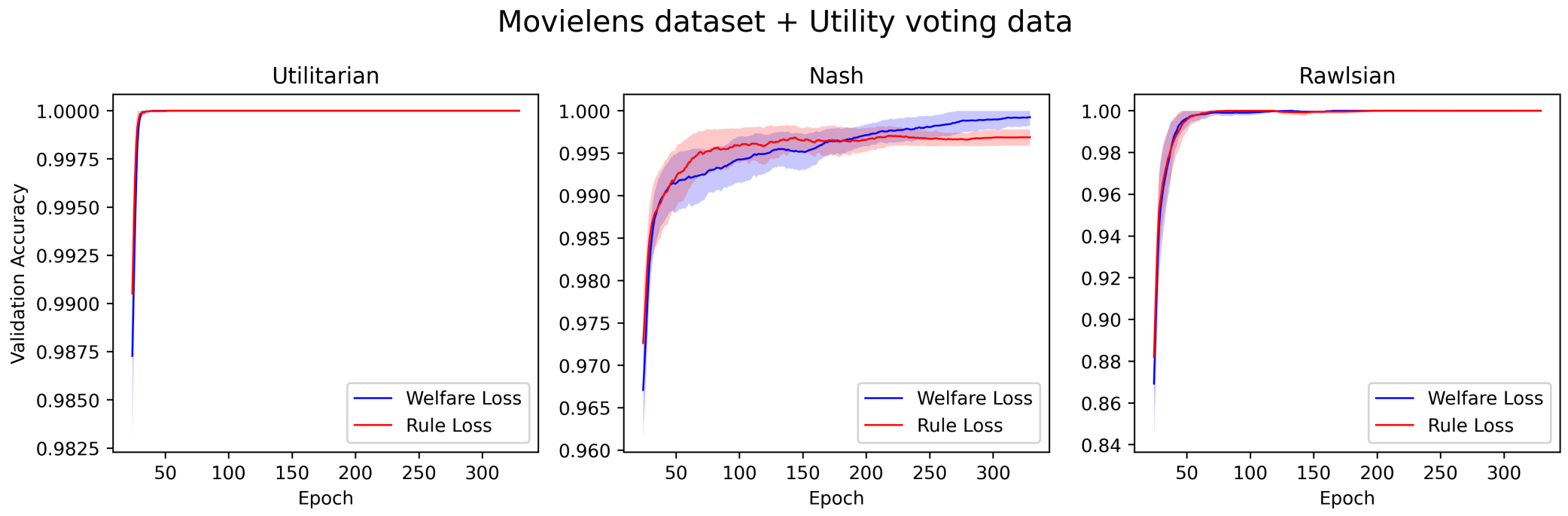

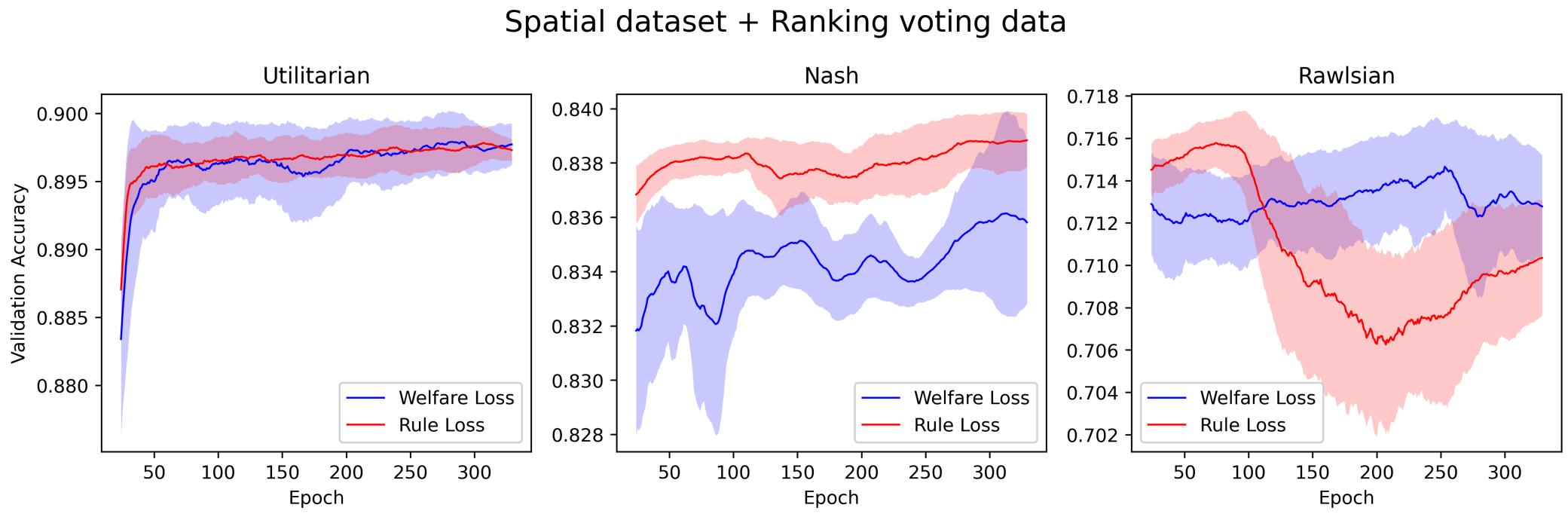

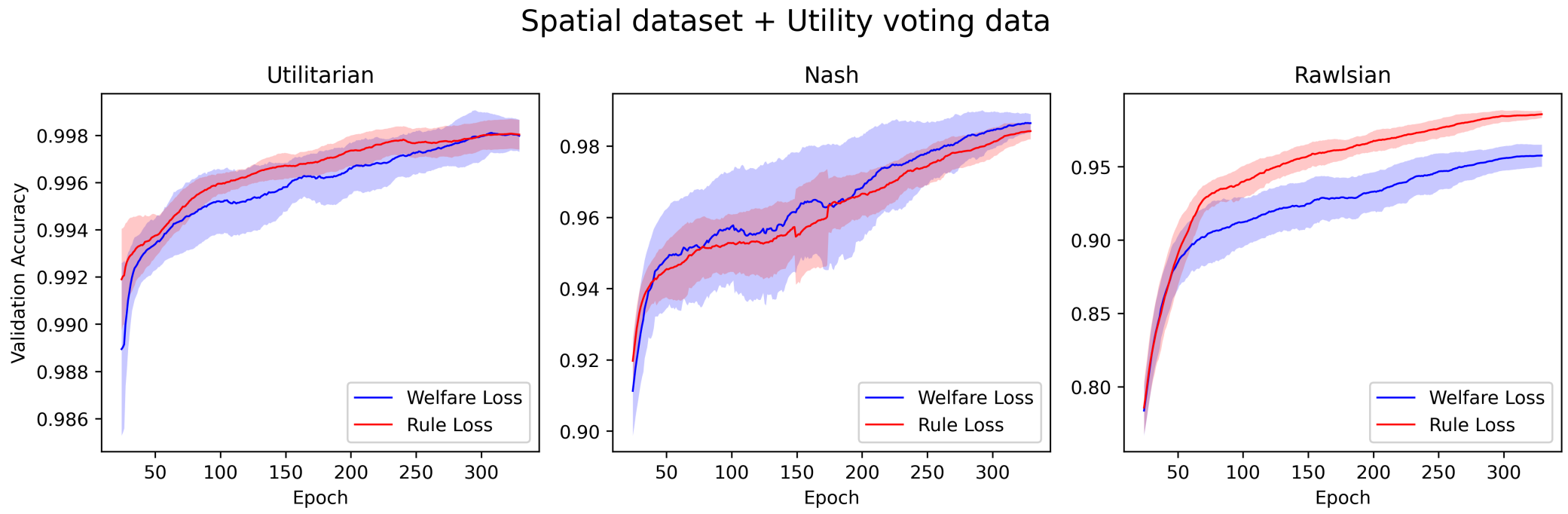

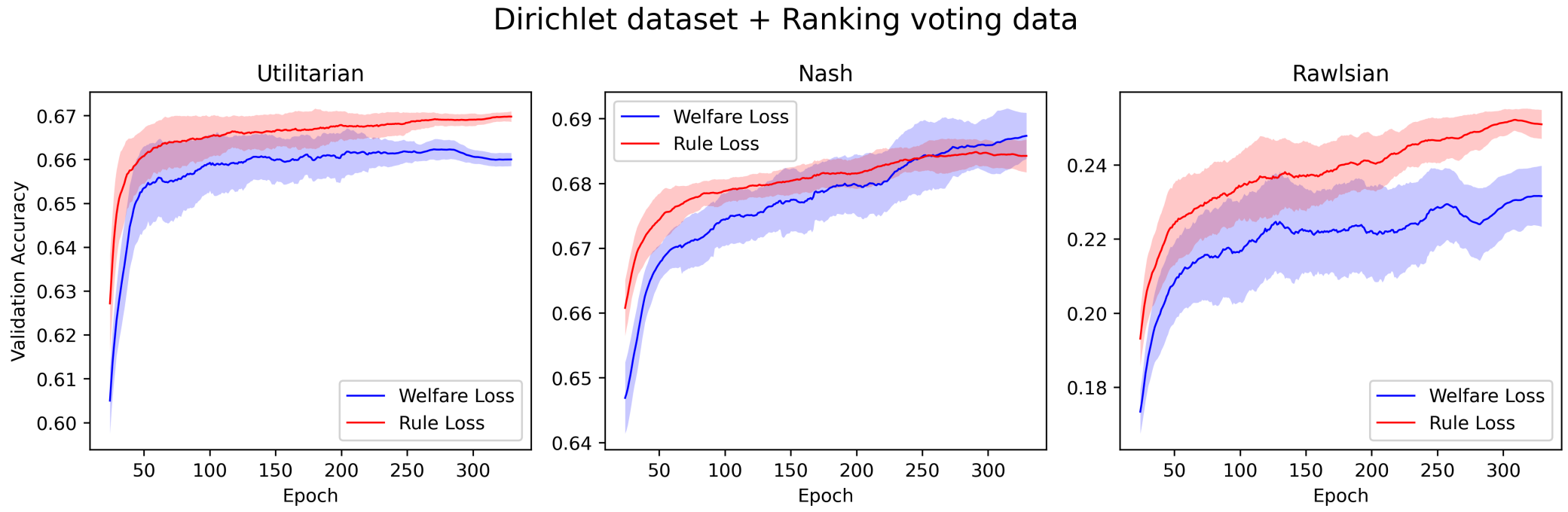

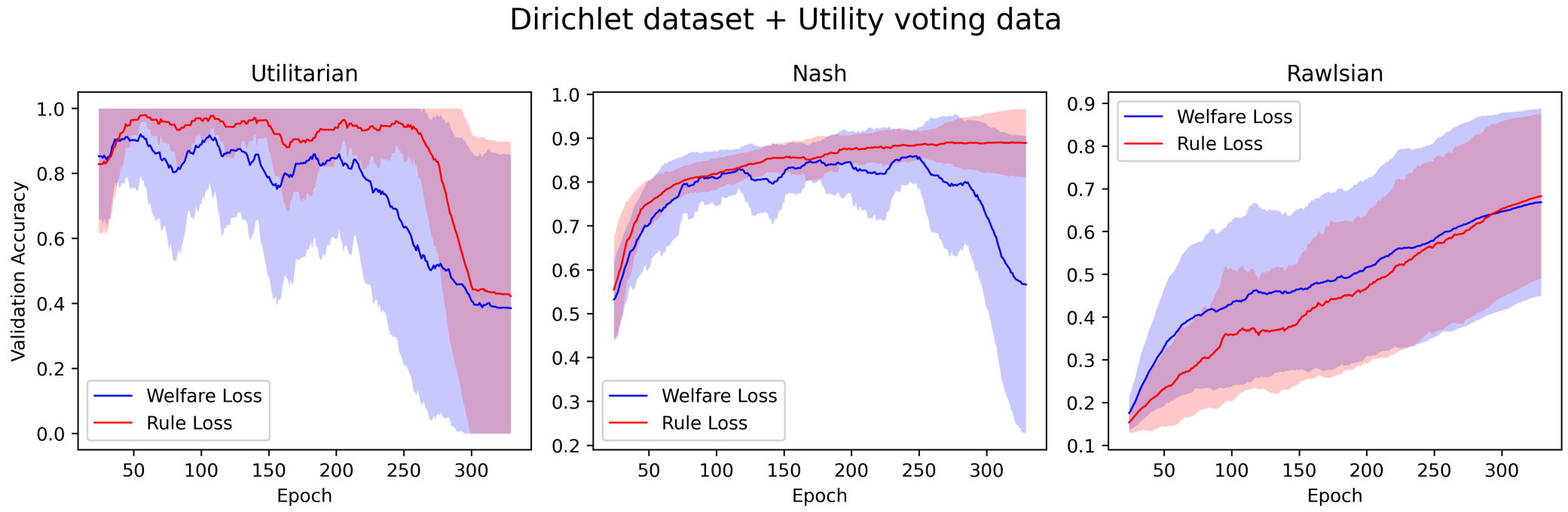

### D.4 Welfare loss closely matches rule loss with known voter utilities

The plots below show that the maximum welfare achieved by both the welfare and the rule loss is the same when the true voter utility data is available in the case of the Movieslens and spatial datasets. The darker line indicates the mean, with the shaded region showing one standard deviation above and below the mean, across five runs, for each plot.

Note that convergence was not achieved by either loss function on the validation set for the Dirichlet dataset (although it was during training). It is likely that this can be improved by further fine-tuning with a larger training set.

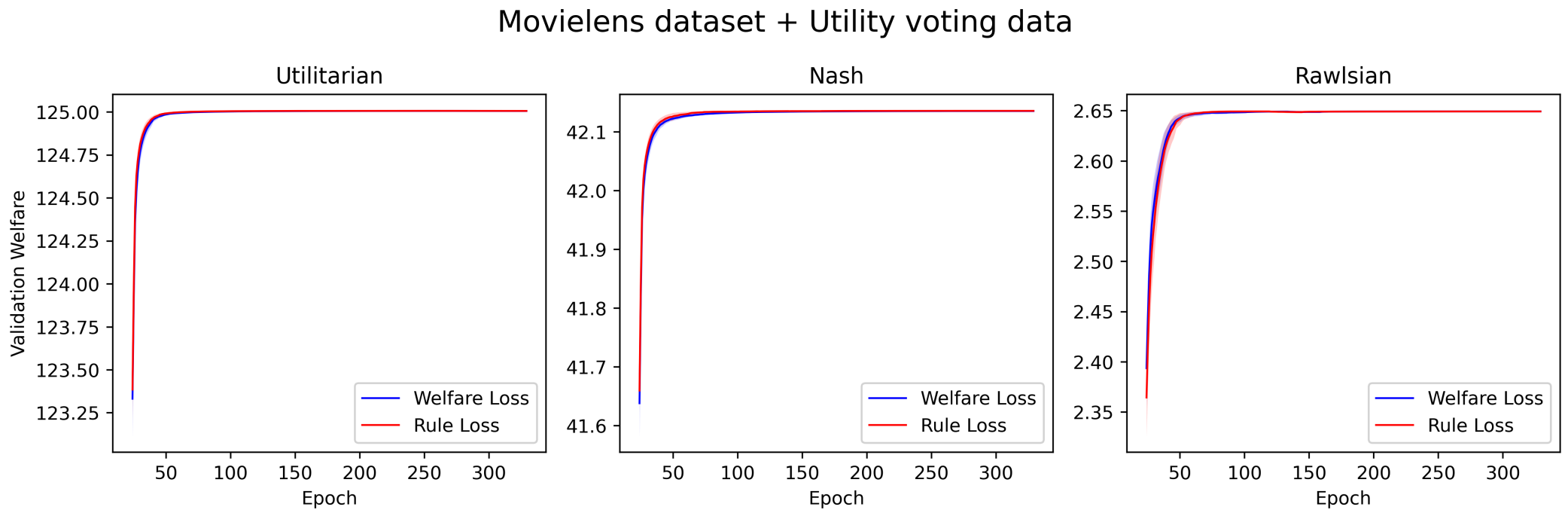

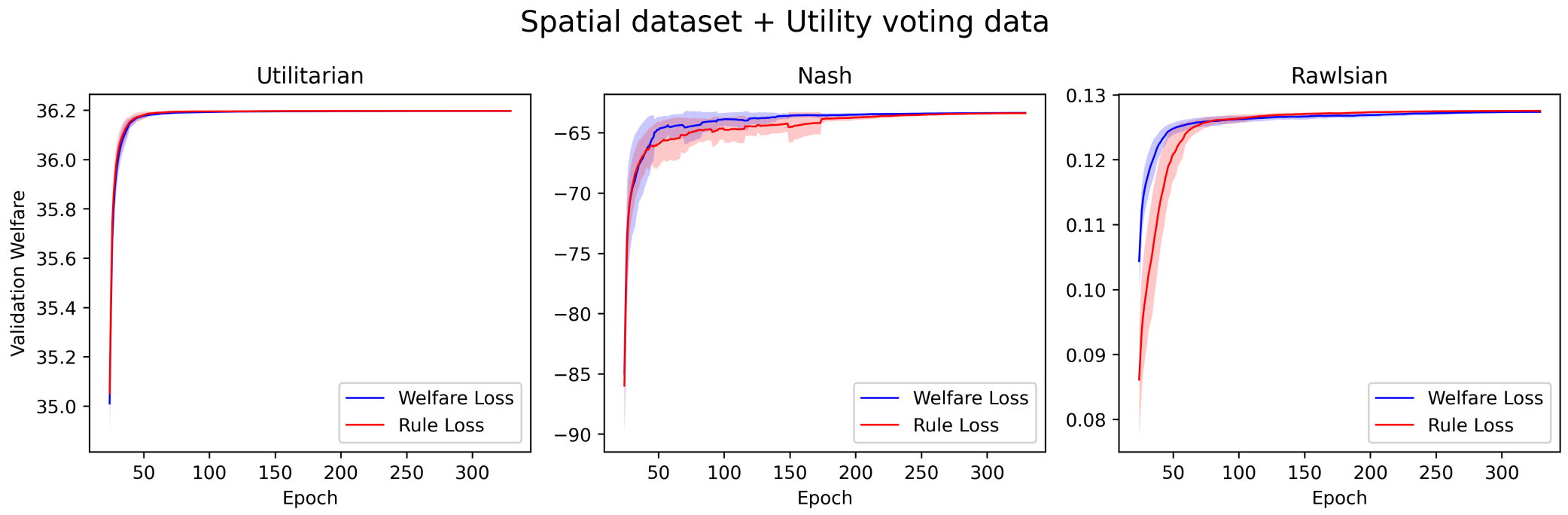

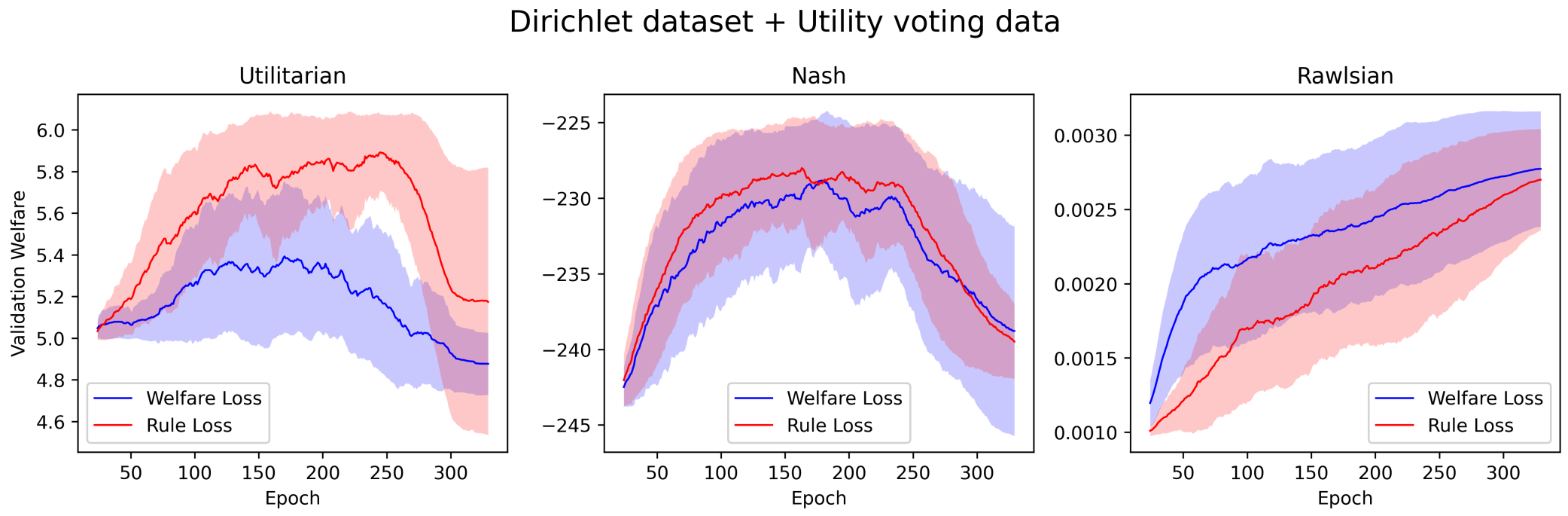

### D.5 Impact of monotonicity loss on learning welfare functions

When enforcing the monotonicity property, through the addition of a monotonicity loss, GEVN is still successful in learning Utilitarian, Nash, and Rawlsian welfare, with no regression in validation or test accuracy (results in Table 3). This is expected since these welfare functions are inherently monotonic, and so successfully learning the welfare function naturally leads to minimizing the added monotonicity loss. Our empirical results provide evidence for the previous statement as the monotonicity loss drops to 0 even when we do not actively enforce the monotonicity constraint.

Table 3: Table showing the average test accuracy and standard deviation on the test set across the five runs.

| **Welfare Function** | **Utilitarian + Mono. Loss** | **Nash + Mono. Loss** | **Rawlsian + Mono. Loss** |
|---|---|---|---|
| **Test Accuracy** | $0.999 \pm 0.001$ | $0.996 \pm 0.001$ | $0.989 \pm 0.006$ |

In the plots below, we see that we are successfully able to reduce the monotonicity loss, without adversely affecting validation accuracy, when learning the Single Transferable Vote rule. This is despite the fact STV is not monotonic in theory — we are able to learn a monotonic local minima.

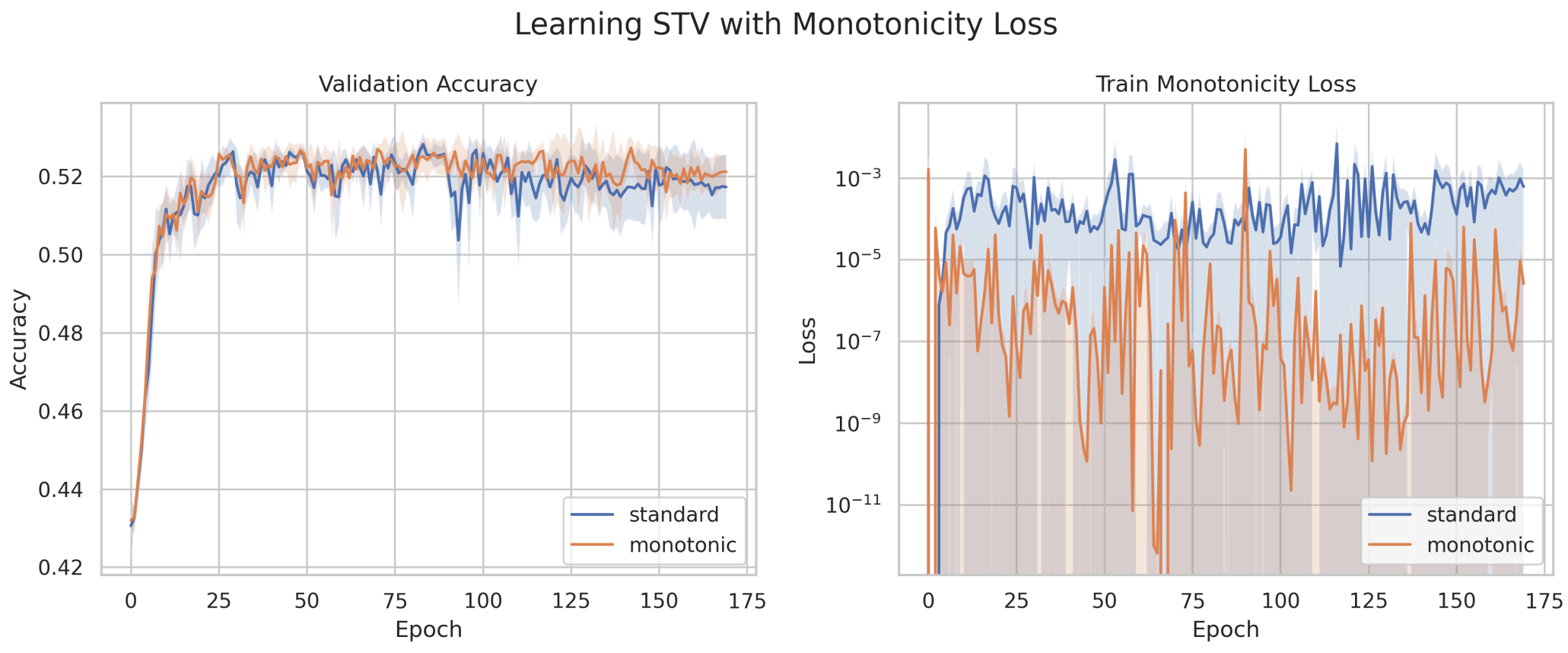

Figure 5: The darker line indicates the mean, with the shaded region showing one standard deviation above and below the mean, across five runs, for each plot.

### D.6 Additional results for adversarial training

The plots below provide the (smoothed using a window of 25) training curves for the adversarial, strategic training process (Section 4.5) using the *private*, *public*, and *results* information settings.

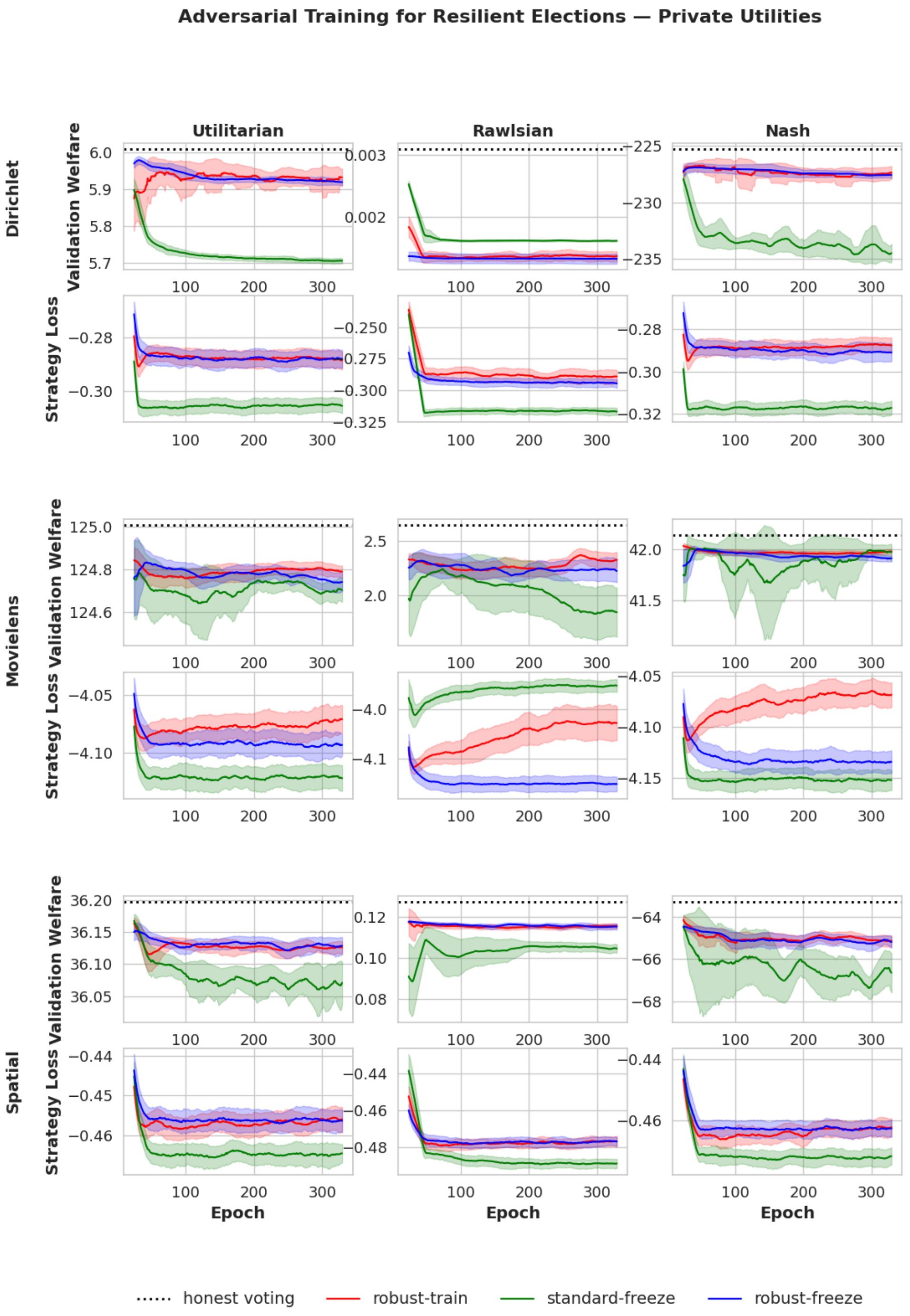

Figure 6: Strategic training in the setting where voters have access to their own utilities. This is a limited information setting, and the voting module with adversarial training, even when frozen, consistently leads to higher social welfare in the presence of 20% strategic voting.

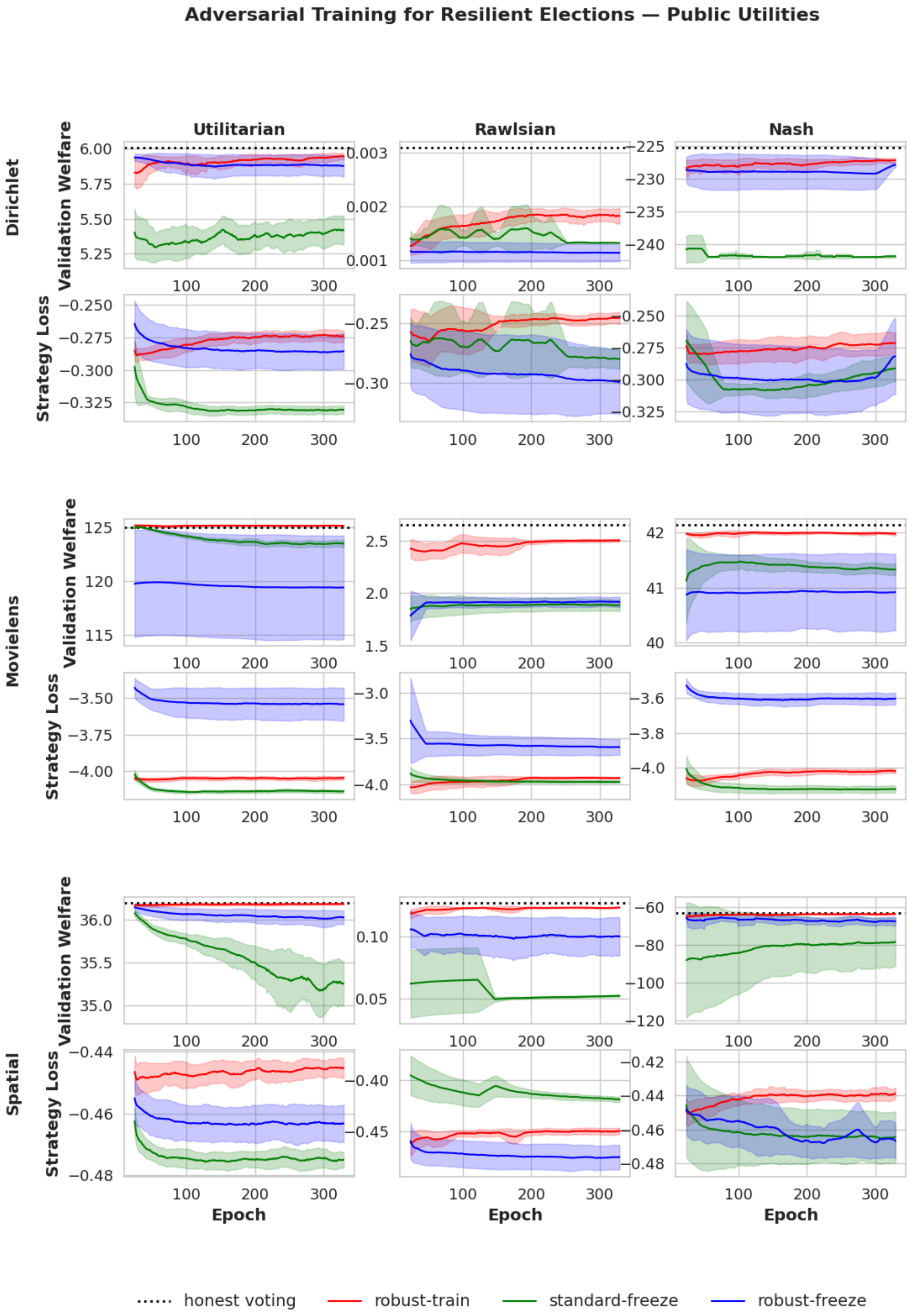

Figure 7: Strategic training in the setting where voters have access to the public utilities of the entire electorate. Although simultaneously training the GEVN and GESN always leads to high social welfare compared to a frozen GEVN, there are combinations of datasets and social welfare functions where the adversarially trained yet frozen GEVN is outperformed by the case without training. That said, note in the majority (6/9) of experimental settings our methodology produces voting methods resilient to strategic voting.

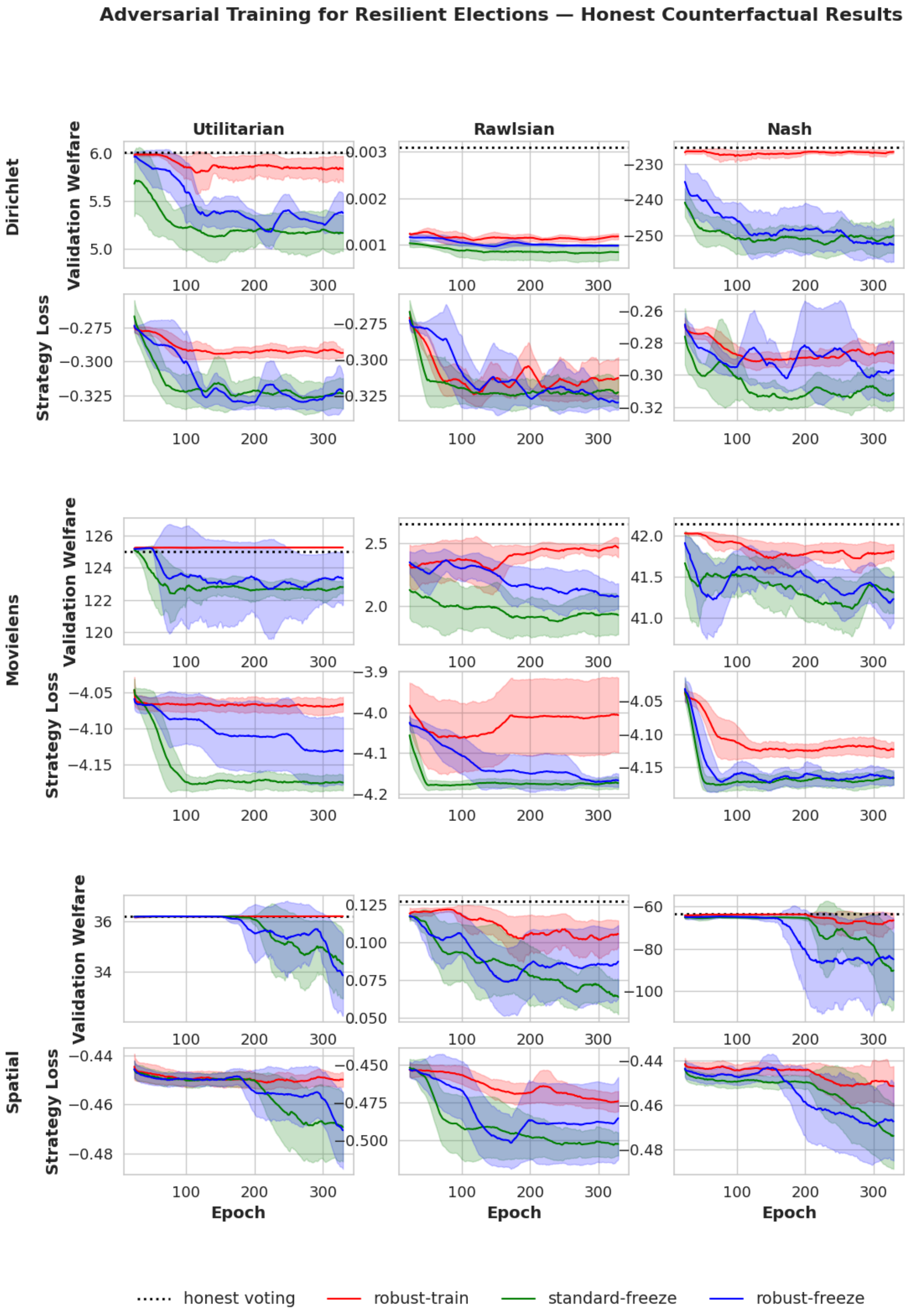

Figure 8: Strategic training in the setting where voters have access to the election results of the hypothetical setting where all voters are honest. Training the GESN simultaneously with strategic voters empirically maintains a stable and high social welfare, and reflects real-world settings where designers are allowed to fine-tune the election mechanism after each election. The GESN is not robust when frozen, reflecting the inherent difficulty of learning robust voting rules against well-informed and motivated actors.

## E Experiment Details and Hyperparameters

This section provides comprehensive details needed to reproduce the experiments. Our methodology is implemented with the PyTorch deep learning framework [41] (BSD License) on top of the PyTorch Geometric library [42] (MIT License), along with NumPy (BSD License), Pandas (BSD License), and tqdm (MIT License). Graphs are generated with Matplotlib (PSF License) and Seaborn (BSD License). Sweeps are handled with Hydra (MIT License) and Pydantic (MIT License). Weights and Biases is used for experiment tracking (MIT License).

### E.1 Datasets

Excluding Table 2, for synthetic datasets, we use training datasets of size $100,000$ with $[3-50]$ voters and $[2-10]$ candidates. Validation datasets are sampled with $1,280$ elements, $75$ voters and $15$ candidates. Test datasets are sampled with $1,280$ elements, $100$ voters and $20$ candidates. Due to dataset limitations, for the Movielens dataset we sample a training dataset of size $5,000$ with $[10-25]$ voters and $[3-7]$ candidates, validation dataset of size $512$ with $30$ voters and $10$ candidates, and test dataset of size $512$ with $45$ voters and $15$ candidates.

**MovieLens**: The MovieLens 32M dataset [33] is a collated collection of 32 million ratings on $87,585$ movies by $200,948$ users. Each movie is rated on a scale of $[0.5, 5]$ in increments of $5$. We filter movies for those with more than $20,000$ ratings, and sample $m$ movies as the candidate set. We then filter users who have expressed a preference for all movies in the candidate set, and then sample $n$ voters. The usage license is found here: https://files.grouplens.org/datasets/movielens/ml-32m-README.html.

**French 2017 Elections (Grenoble) [34]**

This dataset was an online and in situ survey conducted during April 2017, the first round of French elections. We use the data collected from two polling stations in Grenoble, where voters were asked to evaluate each candidate along a continuous line by marking with a pencil. This results in an expressed preference coded from $[0-1]$. The dataset is accessed under the Open Database License.

We speculate preferences collected in this dataset likely contain strategic behaviour as voters preferentially marked candidates at extremes.

### E.2 Network Architecture

**Graph election voting network (GEVN)**

The GEVN builds component MLPs with LayerNorm normalisation [43] and ReLU [44] activation.

- $\psi$ is a 2 layer MLP operating over the concatenation of of edge embeddings with both endpoint node embeddings.
- $\phi_v$ is a 2 layer MLP operating on the concatenation of aggregated messages and the local embedding.
- $\phi_e$ is a 2 layer MLP operating over the concatenation of of edge embeddings with both endpoint node embeddings.

For the small and standard size models respectively, we use node embedding dimensions of $(58, 185)$, edge embedding dimensions of $(19, 60)$, with 4 message passing layers for both. The input and output is transformed to the desired dimension using a linear layer.

**DeepSets election network**

We implement the DeepSet election model using PyTorch Geometric's DeepSetsAggregation utility. The encoder and decoder are MLPs, with 3 and 5 layers and embedding dimensions of 155 and 352 respectively for the small and large model sizes considered. BatchNorm [45] and ReLU are used in the MLPs, and the aggregation function used is sum aggregation.

Table 4: Hyperparameters used during training.

| **Hyperparameter** | **Value** |
|---|---|
| Optimiser | Adam |
| Learning Rate Scheduler (Warmup) | Linear Warmup (Factor $0.1 \rightarrow 1$ across 20 epochs) |
| Learning Rate Scheduler (Main) | Cosine Annealing with Warm Restarts ($T_0 = 20, T_{\text{mult}} = 2$) |
| Learning Rate | 0.0003 |
| Clip Gradient Norm | 1.0 |
| Batch Size | 128 |
| Monotonicity Loss Batch Size | 32 |

**Graph election strategy network (GESN)**

For the *private* information case, we represent the GESN using a custom DeepSet model. The input for $v_i$ is an $\mathbb{R}^m$ dimensional vector representing the $i^{\text{th}}$ row of $\boldsymbol{U}$ — voter information is restricted. In the forward pass of our network, first, the 1 dimensional votes are transformed into 32 dimensional embeddings through a linear layer. Then, in each DeepSet layer, the network passes the embedding through a 2 layer MLP with LeakyReLU [46] activation, followed by concatenation of the sum aggregated embedding across votes, and another 2 layer MLP. We use 2 such DeepSet layers. Finally, the output embeddings are concatenated with the original embeddings as a residual connection, and passed into a 3 layer MLP to obtain the strategic votes. This architecture has $13,729$ parameters in total.

Otherwise, we use a message passing neural network similar to the GEVN.

### E.3 Training Hyperparameters

The second-order differentiation used for monotonicity loss is expensive to compute across all voter-candidate pairs. In training, we subsample candidates in each minibatch to compute the monotonicity loss, and empirically find this provides a sufficiently stable training signal.

Table 4 shows miscellaneous hyper-parameters values used during training:

### E.4 Used Computer Resources

All experiments were run on NVIDIA GPUs.

Experiments on welfare maximisation were performed on an NVIDIA RTX 4080 GPU, with 16GB of VRAM, and 10,240 CUDA cores. The device used an AMD RYZEN 9 7900X CPU with 12 cores and 32GB of RAM. Each individual run takes 1–3 minutes, and we parallelise the execution of runs in sweeps.

Experiments on learning existing voting rules and strategic voting were run on a single NVIDIA RTX 3090 GPU with 24GB of VRAM and 10,496 CUDA cores. The device used an Intel i5-13600KF CPU with 14 cores and 32 GB of RAM. Each individual run takes $3 - 30$ minutes, and we parallelise the execution of runs in sweeps.